\documentclass{article}

\usepackage{amsmath, amssymb}
\usepackage{float}
\usepackage{booktabs}
\usepackage{multirow}
\usepackage{subcaption}
\usepackage{tabularx} 
\usepackage{hyperref}
\usepackage{algorithm}
\usepackage{algpseudocode}
\usepackage{amsmath}
\usepackage{array}
\usepackage{siunitx}
\usepackage[utf8]{inputenc}
\usepackage{booktabs}   
\usepackage{geometry}    
\usepackage{caption}     
\usepackage{rotating}  
\usepackage{graphicx}    
\usepackage{float}      
\usepackage[table]{xcolor}
\usepackage{longtable}
\usepackage{ragged2e}

\title{Enriching Tabular Data with Contextual LLM Embeddings: A Comprehensive Ablation Study for Ensemble Classifiers}

\author{
    Gjergji Kasneci \\
    Technical University of Munich\\
    \texttt{gjergji.kasneci@tum.de} \\
    \&
    Enkelejda Kasneci \\
    Technical University of Munich\\
    \texttt{enkelejda.kasneci@tum.de} \\
}

\begin{document}

\maketitle

\begin{abstract}
    Feature engineering is crucial for optimizing machine learning model performance, particularly in tabular data classification tasks. Leveraging advancements in natural language processing, this study presents a systematic approach to enrich tabular datasets with features derived from large language model embeddings. Through a comprehensive ablation study on diverse datasets, we assess the impact of RoBERTa and GPT-2 embeddings on ensemble classifiers, including Random Forest, XGBoost, and CatBoost. Results indicate that integrating embeddings with traditional numerical and categorical features often enhances predictive performance, especially on datasets with class imbalance or limited features and samples, such as \texttt{UCI Adult}, \texttt{Heart Disease}, \texttt{Titanic}, and \texttt{Pima Indian Diabetes}, with improvements particularly notable in XGBoost and CatBoost classifiers. Additionally, feature importance analysis reveals that LLM-derived features frequently rank among the most impactful for the predictions. This study provides a structured approach to embedding-based feature enrichment and illustrates its benefits in ensemble learning for tabular data.
\end{abstract}

\section{Introduction}

Feature engineering is foundational to machine learning, playing a critical role in determining the performance, interpretability, and scalability of predictive models. Traditional feature engineering methods rely heavily on domain expertise to transform raw data into representations that meaningfully capture relevant patterns. While effective, this approach is often time-intensive and specific to individual applications, limiting its scalability across diverse tasks. The advent of deep learning, particularly in natural language processing (NLP), has introduced powerful new methods for automated feature extraction. These methods, especially contextual embeddings from models like BERT, RoBERTa, and GPT, can capture rich semantic information from textual data, reducing the need for extensive manual preprocessing and enhancing model robustness across domains \cite{devlin2019bert, radford2019language, vaswani2017attention}.

Contextual embeddings offer a valuable means of enhancing tabular datasets by transforming unstructured data into structured, multi-dimensional representations. This transformation has great potential when applied to ensemble classifiers, which have consistently outperformed individual models on a wide range of classification tasks. Ensemble learning methods, such as Random Forest \cite{breiman2001random}, XGBoost \cite{chen2016xgboost}, and CatBoost \cite{dorogush2018catboost}, combine the strengths of multiple learners, making them highly robust and versatile. Given the state-of-the-art performance of these models, integrating advanced embeddings with traditional tabular features could significantly improve classification accuracy and generalizability, creating a hybrid approach that leverages both structured and unstructured data representations.

Moreover, embeddings can serve as proxies for raw data, addressing challenges related to data and model privacy. In sensitive fields like healthcare, finance, and security, embedding techniques enable the extraction of meaningful insights without exposing identifiable information. This approach aligns with the increasing demand for privacy-preserving machine learning, where both data protection and model interpretability are paramount. By using embeddings to represent raw data, this method allows for privacy-friendly feature engineering that retains the value of the original information while minimizing potential privacy risks..

While prior studies have explored the integration of embeddings within various machine learning architectures, most have focused on specific use cases or single classifiers and often lack a systematic approach across multiple datasets and model types \cite{abdelmounaim2024hybrid, al2017using, huang2020tabtransformer, si2019enhancing, sun2019fine}. The effects of embedding-based enrichment on ensemble models across diverse data contexts remain underexplored, leaving a knowledge gap in understanding the generalizability and impact of such techniques. Comprehensive ablation studies, which assess individual contributions of model components, are crucial to address this gap. Ablation analyses can help discern the unique value of embedding-based features relative to baseline features and identify where the combination of embedding and ensemble methods offers the most significant improvements in predictive power.

In this work, we conduct an extensive ablation analysis to assess the impact of embedding-based feature enrichment on ensemble classifiers. Our approach systematically evaluates the integration of embeddings from RoBERTa and GPT-2 into traditional tabular data, comparing performance metrics across several datasets. By examining the performance gains in Random Forest, XGBoost, and CatBoost classifiers, we aim to quantify the benefits of enriched feature representations and provide insights into their contribution to model decisions. Additionally, our feature importance analysis sheds light on the mechanisms and constellations through which contextual embeddings enhance predictive accuracy, thus contributing to a deeper understanding of their role within ensemble learning frameworks. This analysis shows how embedding-based feature enrichment can boost predictive power while also offering a privacy-friendly and efficient approach to modern feature engineering. In summary, this work provides the following contributions:
\begin{enumerate}
    \item We introduce a generic feature enrichment framework that combines contextual embeddings from LLMs with traditional tabular features.

    \item We present a comprehensive ablation study across various benchmark datasets and analyze the performance gains of ensemble classifiers.

    \item By examining the top contributing features, we highlight the importance of embedding-based features and provide a deeper understanding of their influence on model predictions.

    \item  We provide insights into scenarios where embedding-based enrichment is most beneficial, and offer practical guidance on its effective use in classification tasks with structured datasets of varying complexity.

    \item We discuss further potential use cases for embedding-based features, such as privacy-friendly feature representations, reducing reliance on raw, sensitive data without sacrificing predictive power.
\end{enumerate}
The code for this work is open source and available at:\\ \url{https://gitlab.lrz.de/rds/featureenrichmentllmembeddings}.

\section{Related Work}

Feature engineering has been extensively studied, with various approaches proposed to automate and enhance the feature creation process \cite{tsoumakas2010mining, xu2016multi, zheng2018feature}. Traditional methods for feature engineering primarily relied on expert knowledge to craft meaningful representations, but automated feature engineering techniques have emerged to reduce human intervention. Techniques like feature selection, construction, and transformation have been employed to improve model accuracy and interpretability \cite{zheng2018feature}. With the advent of deep learning, feature extraction methods began to incorporate non-linear relationships and hierarchical structures within data, leading to more robust predictive features \cite{bengio2013representation}.

The utilization of deep learning models for feature extraction, particularly in NLP, has gained prominence with models like BERT \cite{devlin2019bert}, GPT-2 \cite{radford2019language}, and their successors, which provide contextual embeddings that capture intricate semantic relationships \cite{raffel2020exploring, wolf2020transformers}. These models leverage transformer-based architectures \cite{vaswani2017attention}, which have revolutionized NLP by enabling self-attention mechanisms that learn contextual dependencies within text sequences. More recent models like RoBERTa \cite{liu2019roberta} and GPT-3 \cite{brown2020language} have further refined these architectures, leading to embeddings that capture even finer-grained semantic nuances. In domains beyond NLP, embeddings derived from transformers have also been applied to structured data, leading to innovations in feature engineering and transfer learning across heterogeneous data types \cite{borisov2022language,cvetkov2023relational,fang2024large,harari2022few,hollmann2022tabpfn,huang2020tabtransformer,nam2024optimized,wang2022transtab}.

Ensemble classifiers have been a staple in machine learning competitions and practical applications due to their robustness and excellent performance on tabular data \cite{borisov2022deep, breiman2001random, chen2016xgboost, dorogush2018catboost, friedman2001greedy}. These classifiers combine multiple models to improve generalization and reduce variance, with methods like bagging, boosting, and stacking being widely adopted. In recent years, researchers have explored hybrid models that combine traditional ensemble techniques with neural networks to leverage both structured data and unstructured embeddings, yielding significant performance gains \cite{fort2019deep, he2017neural, rendle2020neural}. Studies have shown that combining traditional features with deep learning-based embeddings can lead to significant performance gains, especially in domains where feature interactions are complex and multi-modal data is involved \cite{abdelmounaim2024hybrid, al2017using, huang2020tabtransformer, minaee2021deep, si2019enhancing, sun2019fine}.

In parallel, approaches for integrating embeddings into ensemble classifiers have been evaluated for their ability to capture hidden patterns within high-dimensional data \cite{abdelmounaim2024hybrid,  al2017using, zhou2019deep}. For instance, embedding-based feature engineering has proven effective in applications like sentiment analysis, recommendation systems, and medical diagnostics, where context-rich representations enhance model interpretability and predictive power \cite{si2019enhancing, wu2020deep}. Recently, methods like knowledge distillation and transfer learning have been employed to integrate pre-trained embeddings, enhancing the adaptability of ensemble classifiers across tasks and datasets \cite{cai2020transfer, gou2021knowledge, theodoris2023transfer}.

Ablation studies are essential for understanding the contribution of individual components within a machine learning pipeline \cite{hooker2019benchmarking, meyes2019ablation, sundararajan2017axiomatic}. They help in identifying which features or components significantly impact the model's performance, thereby informing feature selection and model optimization strategies. However, comprehensive ablation analyses that explore the integration of contextual embeddings with ensemble classifiers across diverse datasets remain limited. Prior work has primarily focused on specific use cases or individual classifiers, without a systematic ablation approach to assess the predictive quality of feature subsets across multiple models and datasets \cite{shi2020safe}. Furthermore, recent advances in explainable AI emphasize the importance of understanding how each feature subset contributes to model predictions, motivating more granular analyses of feature interactions in complex models \cite{lipton2018mythos, molnar2020interpretable}.

This work addresses the above gaps by providing a systematic evaluation of embedding-based feature enrichment in ensemble learning frameworks, thereby contributing to a deeper understanding of feature integration strategies in machine learning. By examining the combined effects of baseline and embedding-enriched features across multiple datasets, this work provides insights into the conditions under which embedding-based feature enrichment enhances model performance, complementing current research in both feature engineering and ensemble learning.

\section{Feature Enrichment Framework for LLM Embeddings}

Feature enrichment in machine learning involves augmenting the existing feature set with additional features to enhance model performance. This process may integrate both structured and unstructured data sources to enrich the feature space and enable models to capture more complex patterns. In this work, we implement a feature enrichment strategy that combines baseline structured features with embeddings derived from pre-trained language models. To ensure computational efficiency and remain as parsimonious as possible in the enrichment process, we apply Principal Component Analysis (PCA) for initial dimensionality reduction and noise reduction on the embedding dimensions, followed by feature selection to retain only the most informative embedding dimensions.

\subsection{General Framework and Definitions}

We begin by defining the fundamental concepts and notations used throughout this section.

\begin{itemize}
    \item Let the \textbf{tabular dataset} be represented as \( \mathcal{D} = \{(\mathbf{x}_i, y_i)\}_{i=1}^n \), where \( n \) is the number of samples. Here, \( \mathbf{x}_i \in \mathbb{R}^p \) denotes the feature vector for the \( i \)-th sample, and \( y_i \) is the corresponding target label.
    \item The \textbf{baseline feature matrix} \( \mathbf{X}_{\text{baseline}} \in \mathbb{R}^{n \times p} \) encompasses all original features for each sample in the dataset.
    \item A set of \( K \) \textbf{pre-trained language models}, \( \mathcal{M} = \{ M_1, M_2, \dots, M_K \} \), serves as \textbf{feature generators}. For instance, \( M_1 \) may correspond to GPT-2, and \( M_2 \) to RoBERTa.
    \item The \textbf{transformation function} \( \phi: \mathbb{R}^p \rightarrow \mathcal{T} \) converts structured baseline features into textual representations suitable for embedding generation, where \( \mathcal{T} \) denotes the space of textual data.
    \item For each language model \( M_k \in \mathcal{M} \), an \textbf{embedding function} \( f_{M_k}: \mathcal{T} \rightarrow \mathbb{R}^{d_k} \) maps textual input to a \( d_k \)-dimensional embedding vector.
\end{itemize}

\subsection{Feature Enrichment Process}

The feature enrichment process consists of three primary stages: (1) transformation of structured data into textual format, (2) embedding generation and initial dimensionality reduction using PCA, and (3) feature selection to retain the most informative embedding dimensions.

\textbf{Stage 1: Transformation to Textual Representation}

This stage involves converting structured data into a textual format suitable for input into pre-trained language models~\cite{borisov2022language}.

\begin{enumerate}
    \item For each feature vector \( \mathbf{x}_i = [x_{i1}, x_{i2}, \dots, x_{ip}] \), we apply the transformation function \( \phi \) to generate a textual representation:
    \[
    \phi(\mathbf{x}_i) = \text{\small ``}\texttt{Feature1\_name:} \; x_{i1}, \; \texttt{Feature2\_name:} \; x_{i2}, \; \dots, \; \texttt{FeatureP\_name:} \; x_{ip}\text{\small ''}
    \]
    \item We construct a corpus of textual representations for the entire dataset:
    \[
    \mathbf{X}_{\text{text}} = \{\phi(\mathbf{x}_1), \dots, \phi(\mathbf{x}_n)\}
    \]
\end{enumerate}

\textbf{Stage 2: Embedding Generation and PCA Dimensionality Reduction}

Given the high dimensionality of embeddings from advanced language models, we apply PCA to remove noise and reduce the embeddings to a manageable size before feature selection.

\begin{enumerate}
    \setcounter{enumi}{2}
    \item For each language model \( M_k \in \mathcal{M} \), we generate an initial embedding matrix \( \mathbf{E}_{M_k} \in \mathbb{R}^{n \times D_k} \) by applying the embedding function \( f_{M_k} \) to the textual corpus:
    \[
    \mathbf{E}_{M_k} = f_{M_k}(\mathbf{X}_{\text{text}})
    \]
    where \( D_k \) is the original embedding dimension (e.g., 768 for GPT-2).
    \item We apply PCA to reduce each embedding matrix \( \mathbf{E}_{M_k} \) to a lower dimension \( d \) (e.g., \( d = 50 \)), resulting in a reduced embedding matrix \( \mathbf{E}_{M_k}^{\text{PCA}} \in \mathbb{R}^{n \times d} \):
    \[
    \mathbf{E}_{M_k}^{\text{PCA}} = \text{PCA}(\mathbf{E}_{M_k}, \text{n\_components}=d)
    \]
\end{enumerate}

\textbf{Stage 3: Feature Selection via Random Forest Importance}

To further enhance computational efficiency and focus on the most informative features, we perform feature selection on the PCA-reduced embeddings.

\begin{enumerate}
    \setcounter{enumi}{4}
    \item We train a Random Forest classifier \( C_k \) using \( \mathbf{E}_{M_k}^{\text{PCA}} \) to estimate feature importance scores for each embedding dimension.
    \item Based on the importance scores, we select the top \( m \) most informative dimensions from each \( \mathbf{E}_{M_k}^{\text{PCA}} \). Let \( \mathcal{J}_{M_k} \subset \{1, \dots, d\} \) denote the indices of these top dimensions. The selected embedding matrix is then:
    \[
    \mathbf{E}_{M_k}^{\text{selected}} = \mathbf{E}_{M_k}^{\text{PCA}}[:, \mathcal{J}_{M_k}]
    \]
\end{enumerate}

\textbf{Final Enriched Feature Matrix}

We construct the final enriched feature matrix \( \mathbf{F}_{\text{enriched}} \) by concatenating the baseline features with the selected embedding dimensions from all language models:

\[
\mathbf{F}_{\text{enriched}} = \left[ \mathbf{X}_{\text{baseline}} \, \vert \, \mathbf{E}_{M_1}^{\text{selected}} \, \vert \, \dots \, \vert \, \mathbf{E}_{M_K}^{\text{selected}} \right]
\]

This comprehensive approach ensures that we integrate rich contextual information from language models while maintaining computational tractability and model interpretability.

\section{Methodology for Classifier Evaluation and Ablation Study}
\label{sec:methodology}

This section presents the methodological framework used to assess the efficacy of embedding-based feature enrichment in enhancing ensemble classifiers. We detail the processes of dataset preparation, feature enrichment, classifier training, ablation study design, evaluation metrics, and statistical validation to ensure the robustness and reproducibility of our findings.

\subsection{Experimental Setup}
In the following, we describe the experimental setup developed to accurately evaluate the effects of embedding-based feature enrichment on classification performance. This approach encompasses dataset selection, preprocessing, feature engineering, and classification model configuration to ensure a systematic and reproducible evaluation process.

\begin{table}[H]
    \scriptsize
    \centering
    \caption{Comprehensive Overview of Datasets Used in the Study}
    \label{tab:dataset_overview}
    \renewcommand{\arraystretch}{1.2}
    \setlength{\tabcolsep}{4pt}
    
    \begin{tabularx}{\textwidth}{ 
        >{\raggedright\arraybackslash}X
        >{\centering\arraybackslash}c
        >{\centering\arraybackslash}c
        >{\centering\arraybackslash}c
        >{\centering\arraybackslash}c
        >{\centering\arraybackslash}c
        >{\raggedright\arraybackslash}X
        >{\centering\arraybackslash}c
        >{\centering\arraybackslash}c
    }
        \toprule
        \textbf{Dataset} & \textbf{Classes} & \textbf{Instances} & \textbf{Tot. Feat.} & \textbf{Cat. Feat.} & \textbf{Num. Feat.} & \textbf{Class Dist.} & \textbf{Miss. Vals.} & \textbf{Source} \\
        \midrule
        \rowcolor{gray!10}
        \textbf{\texttt{UCI Adult}} & 2 & 65,123 & 14 & 8 & 6 & 24\% $>$\$50K, 76\% $\leq$\$50K & Yes & UCI \\
        \addlinespace
        \rowcolor{white}
        \textbf{\texttt{UCI Heart Disease}} & 2 & 303 & 13 & 5 & 8 & 54\% No Disease, 46\% Disease & Yes & UCI \\
        \addlinespace
        \rowcolor{gray!10}
        \textbf{\texttt{UCI Wine Quality}} & 10 & 1,599 & 11 & 0 & 11 & Varies Across 10 Classes & No & UCI \\
        \addlinespace
        \rowcolor{white}
        \textbf{\texttt{Titanic}} & 2 & 1,309 & 12 & 3 & 4 & 38\% Survived, 62\% Not Survived & Yes & Kaggle \\
        \addlinespace
        \rowcolor{gray!10}
        \textbf{\texttt{Pima Indians Diabetes}} & 2 & 768 & 8 & 0 & 8 & 35\% Positive, 65\% Negative & No & UCI \\
        \addlinespace
        \rowcolor{white}
        \textbf{\texttt{Breast Cancer Wisconsin}} & 2 & 569 & 30 & 0 & 30 & 63\% Benign, 37\% Malignant & No & UCI \\
        \addlinespace
        \rowcolor{gray!10}
        \textbf{\texttt{Car Evaluation}} & 4 & 1,728 & 6 & 6 & 0 & 70\% Unacc., 22\% Acc., 4\% Good, 4\% Vgood & No & UCI \\
        \addlinespace
        \rowcolor{white}
        \textbf{\texttt{UCI Letter Recognition}} & 26 & 20,000 & 17 & 0 & 16 & Approximately Uniform & No & UCI \\
        \addlinespace
        \rowcolor{gray!10}
        \textbf{\texttt{UCI Covertype}}$^*$ & 7 & 581,012 & 54 & 44 & 10 & Varies Across 7 Classes & No & UCI \\
        \bottomrule
    \end{tabularx}
\end{table}

\paragraph{Datasets.} To ensure the generalizability of our results across diverse domains, we utilized ten benchmark datasets of varying sizes  and complexities (in terms of features and instances) from the UCI Machine Learning Repository\footnote{https://archive.ics.uci.edu/} and Kaggle\footnote{https://www.kaggle.com/datasets}. The selected datasets encompass binary and multi-class classification tasks with differing types and numbers of features and instances. Table~\ref{tab:dataset_overview} provides a comprehensive overview of the relevant aspects of these datasets.

$^*$Except for the \texttt{UCI Covertype} dataset -- where 15\% of the data was used, sampled through stratified sampling -- all other datasets were used in their full form.

\paragraph{Data Preprocessing.} Each dataset underwent careful and rigorous preprocessing to ensure data quality and suitability for analysis:

\begin{itemize}
    \item \textbf{Handling Missing Values}: Missing values in categorical features were imputed using the mode while missing numerical values were imputed with the mean. Rows with missing target values (i.e., class labels) were dropped.
    
    \item \textbf{Encoding Categorical Variables}: One-Hot Encoding was utilized for nominal categorical variables, transforming them into binary vectors.
    
    \item \textbf{Feature Scaling}: Numerical features were standardized using the \textit{StandardScaler} to ensure zero mean and unit variance.
    
    \item \textbf{Target Variable Encoding}:
    \begin{itemize}
        \item For \textbf{binary classification} tasks, target labels were mapped to 0 and 1.
        \item For \textbf{multi-class classification} tasks, the target variable was encoded using \textit{LabelEncoder}.
    \end{itemize}
\end{itemize}

\subsection{Feature Engineering and Enrichment}

\paragraph{Embedding Generation and Dimensionality Reduction.} We generated contextual embeddings using two pre-trained language models, namely RoBERTa and GPT-2. The embedding generation and dimensionality reduction process is formalized in three steps as follows:

\begin{enumerate}
    \item The step of \textbf{textual transformation}, where the following transformation is applied to each instance \(\mathbf{x}_i\):
    \[
    \phi(\mathbf{x}_i) = \text{Concatenate}\left( \{ f_j\!:\! v_{ij} \}_{j=1}^p \right)
    \]
    where \( f_j \) are feature names and \( v_{ij} \) are feature values.

    \item The step of \textbf{embedding generation}, where, for each pre-trained language model $M_k$, we map the textual representation of the tabular dataset into the embedding space to obtain the corresponding embedding matrix:
    \[
    \mathbf{X}_{\text{text}} = \{\phi(\mathbf{x}_1), \dots, \phi(\mathbf{x}_n)\}
    \]
    \[
    \mathbf{E}_{M_k} = f_{M_k}(\mathbf{X}_{\text{text}})
    \]
    where \( f_{M_k} \) is the embedding function for model \( M_k \).

    \item The step of noise removal through \textbf{PCA-based dimensionality reduction}:
    \[
    \mathbf{E}_{M_k}^{\text{PCA}} = \text{PCA}(\mathbf{E}_{M_k}, \text{n\_components}=d)
    \]
    with \( d = 50 \) to still have a high coverage of relevant dimensions.
\end{enumerate}

\paragraph{Feature Selection.} We employed Random Forest feature importance to select and retain the most informative embedding dimensions: 

\begin{enumerate}
    \item We conduct a \textbf{feature importance estimation} by applying a Random Forest classifier $C_k$ to $\mathbf{E}_{M_k}^{\text{PCA}}$. Specifically, we train $C_k$ on $\mathbf{E}_{M_k}^{\text{PCA}}$ to obtain importance scores.
   
    \item Then we \textbf{select the top features} based on the computed importance score as:
    \[
    \mathbf{E}_{M_k}^{\text{selected}} = \mathbf{E}_{M_k}^{\text{PCA}}[:, \mathcal{J}_{M_k}]
    \]
    where \( \mathcal{J}_{M_k} \) are indices of the top \( m = 10 \) features. Importantly, our analysis tested different numbers of features from the PCA-reduced embeddings. We found that selecting the top 10 features captured all the relevant information for the classification tasks. This approach remained efficient and parsimonious, balancing accuracy with simplicity.
\end{enumerate}

\subsection{Classifier Selection and Training}

\paragraph{Classifiers.} We used three ensemble classifiers, each known for robustness and strong performance on tabular data~\cite{borisov2022deep}:

\begin{enumerate}
    \item \textbf{Random Forest Classifier} \cite{breiman2001random}: An ensemble of decision trees utilizing bagging to improve generalization and reduce overfitting.
    \item \textbf{XGBoost Classifier} \cite{chen2016xgboost}: An optimized gradient boosting framework known for its efficiency and high performance.
    \item \textbf{CatBoost Classifier} \cite{dorogush2018catboost}: A gradient boosting library that inherently handles categorical features and employs ordered boosting to mitigate overfitting.
\end{enumerate}

\paragraph{Hyperparameter settings.}
In this study, classifiers were configured with consistent hyperparameters across all feature subsets and datasets to attribute performance improvements directly to the impact of embedding-based feature enrichment rather than model-specific tuning. By standardizing these settings, we ensure a fair comparison that isolates the contributions of enriched features, highlighting their intrinsic value in boosting model accuracy and feature importance. For this purpose, we used the following hyperparameters: Random Forest with 100 estimators and balanced class weights; XGBoost with default tree depth, learning rate of 0.1, and ‘mlogloss’ as the evaluation metric; and CatBoost with default parameters and a random seed for reproducibility.

\subsection{Ablation Study Design}
To thoroughly assess the contribution of embedding-based feature enrichment, we designed a structured ablation study that isolates the effects of various feature subsets on classifier performance. This approach enables a clear comparison between models with baseline features and those enhanced by RoBERTa and GPT-2 embeddings, quantifying their impact on predictive accuracy.

\paragraph{Feature Subset Definitions.} For our ablation study, we defined the following feature subsets to pinpoint which combinations of baseline and embedding features yield the highest predictive value:

\begin{enumerate}
    \item \textbf{Baseline}: Original preprocessed features.
    \item \textbf{GPT2\_Selected}: Selected GPT-2 embedding features.
    \item \textbf{RoBERTa\_Selected}: Selected RoBERTa embedding features.
    \item \textbf{Baseline\_GPT2\_Selected}: Baseline features with selected GPT-2 embeddings.
    \item \textbf{Baseline\_RoBERTa\_Selected}: Baseline features with selected RoBERTa embeddings.
    \item \textbf{GPT2\_RoBERTa\_Selected}: Combined selected embeddings from both models.
    \item \textbf{Baseline\_GPT2\_RoBERTa\_Selected}: Baseline features with selected embeddings from both models.
\end{enumerate}

\paragraph{Evaluation Strategy for the Ablation Study.} The ablation study evaluates the performance of the classifiers on the above feature subsets with the goal to assess the contribution of embedding-based features. For each dataset and classifier:

\begin{enumerate}
    \item For the validation of the models, we perform  a stratified \textbf{5-fold cross-validation} is performed. Specifically, each classifier is trained on the training folds for each feature subset.
    \item Relevant \textbf{classification performance metrics} such as Accuracy, Balanced Accuracy, Weighted F1 Score, and ROC-AUC are computed on the validation fold.
    \item \textbf{Paired t-tests} compare performance across feature subsets.
\end{enumerate}

\paragraph{Algorithmic Procedure.} For clarity and reproducibility, the ablation study procedure is encapsulated in Algorithm~\ref{alg:ablation_study}. It outlines each step of our approach for evaluating feature subset contributions to classifier performance.

\begin{figure}[h!]
\centering
\makebox[\textwidth]{ 
    \includegraphics[width=1.15\textwidth]{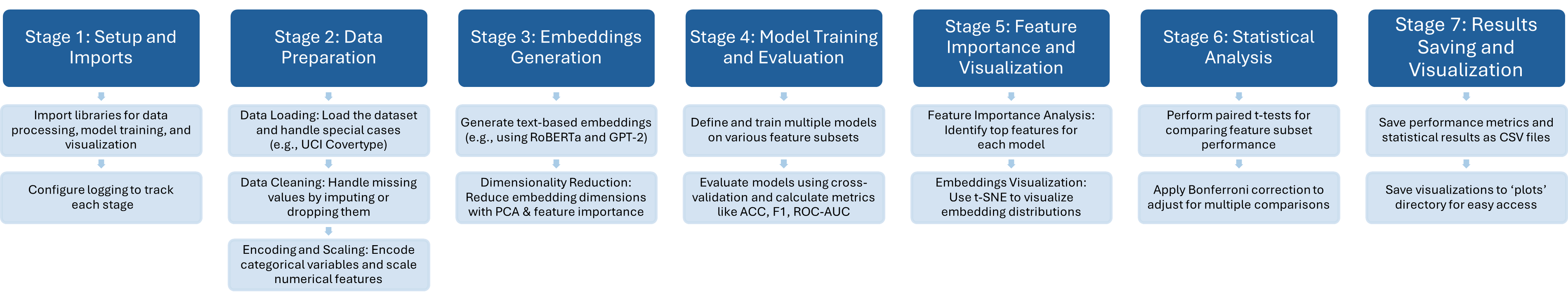}
}
\caption{Overview of the experimental workflow stages.}
\label{fig:pipeline}
\end{figure}

In addition to the robust and systematic preprocessing, feature subset selection, and embedding pipeline, Algorithm~\ref{alg:ablation_study} employs stratified cross-validation and paired statistical testing to provide robust, comparative insights into the predictive power of enriched features. This modular, step-by-step approach enables the systematic isolation of feature subset effects, enhancing the interpretability and precision of the ablation study’s results. Figure~\ref{fig:pipeline} depicts the pipeline used in the study.

\begin{algorithm}[H]
\caption{Ablation Study Procedure}
\label{alg:ablation_study}
\begin{algorithmic}[1]
\Require Dataset \( D \), Feature Subsets \( \mathcal{F} \), Classifiers \( \mathcal{C} \)
\For{each dataset \( D \in \mathcal{D} \)}
    \State Preprocess dataset \( D \)
    \State Encode categorical variables and scale and normalize numerical features
    \State Generate textual representations from \( D \)
    \State Create embeddings using transformer models (e.g., RoBERTa, GPT-2)
    \State Apply PCA to reduce embedding dimensions to $d=50$
    \State Select top-10 features based on feature importance and apply t-SNE for 2D visualization
    \For{each classifier \( C \in \mathcal{C} \)}
        \For{each feature subset \( F \in \mathcal{F} \)}
            \State Initialize lists to store Accuracy, Balanced Accuracy, Weighted F1 Score, ROC-AUC
            \State Create 5 stratified folds for cross-validation from dataset \( D \)
            \For{fold \( k = 1 \) to \( 5 \)}
                \State \textbf{Define} fold \( k \) as the validation set \( (X_{\text{val}}^k, y_{\text{val}}^k) \)
                \State \textbf{Define} the remaining folds as the training set \( (X_{\text{train}}^k, y_{\text{train}}^k) \)
                \State Train classifier \( C \) on \( X_{\text{train}}^k, y_{\text{train}}^k \) using features from subset \( F \)
                \State Predict labels \( \hat{y}^k \) using \( C \) on \( X_{\text{val}}^k \)
                \State Compute Accuracy, Balanced Accuracy, Weighted F1 Score, ROC-AUC
                \State Record the computed metrics for fold \( k \)
            \EndFor
            \State Compute average of each metric over the 5 folds for classifier \( C \) and feature subset \( F \)
            \State \textbf{Analyze} feature importance using classifier \( C \) on subset \( F \)
            \State \textbf{Select} top-10 most important features based on feature importance scores
            \State \textbf{Save} results for metrics, feature importance
        \EndFor
        \State Perform paired t-tests to compare metrics across feature subsets \( \mathcal{F} \) for \( C \) and save results
    \EndFor
\EndFor
\end{algorithmic}
\end{algorithm}

\subsection{Evaluation Metrics}

To comprehensively assess classifier performance and capture various aspects of predictive accuracy, we employed multiple evaluation metrics, ensuring a robust and nuanced evaluation:

\begin{itemize}
    \item We use \textbf{Accuracy} and \textbf{Balanced Accuracy} to assess general performance and the ability to handle class imbalance, respectively:
    \[
    \text{Accuracy} = \frac{1}{n} \sum_{i=1}^n \mathbf{1}\{\hat{y}_i = y_i\}
    \]
    where \( n \) is the total number of instances, \( \hat{y}_i \) is the predicted label, and \( y_i \) is the true label for instance \( i \).

    \[
    \text{Balanced Accuracy} = \frac{1}{C} \sum_{c=1}^C \frac{\text{TP}_c}{\text{TP}_c + \text{FN}_c}
    \]
    where \( C \) is the number of classes, \( \text{TP}_c \) and \( \text{FN}_c \) are the true positives and false negatives for class \( c \), respectively. Balanced Accuracy accounts for class imbalance by averaging recall over all classes.

    \item To analyze the precision and recall-based performance, we include the \textbf{weighted F1 score}, which is particularly important in datasets with class imbalance:
    \[
    \text{Weighted F1 Score} = \sum_{c=1}^C \frac{N_c}{N} \cdot \text{F1 Score}_c
    \]
    where \( N_c \) is the number of true instances in class \( c \), \( N \) is the total number of instances, and \( \text{F1 Score}_c \) is the F1 Score for class \( c \). The F1 Score for each class is calculated as:
    \[
    \text{F1 Score}_c = 2 \cdot \frac{\text{Precision}_c \cdot \text{Recall}_c}{\text{Precision}_c + \text{Recall}_c}
    \]
    with precision and recall defined for each class.

    \item To analyze the discriminative performance through a threshold-independent performance measure, we employ \textbf{ROC-AUC}:
    \[
    \text{ROC-AUC} = \int_{0}^{1} \text{TPR}(FPR^{-1}(t)) \, dt
    \]
    For multi-class classification, we use the macro-averaged ROC-AUC computed using the One-vs-Rest approach. In addition, all ROC curves were plotted for visual assessment and comparisons.

    \item For fine-granular per-class performance analysis, we generate a \textbf{classification report} with detailed per-class metrics, including all the metrics and the paired t-test introduced above.
\end{itemize}

\subsection{Statistical Significance Testing}

Paired t-tests were conducted between all pairs of feature subsets for each classifier and dataset combination. The \textbf{Bonferroni correction} was applied to adjust for multiple comparisons, ensuring thus a family-wise significance level of \( \alpha = 0.05 \), which overall is way more conservative than using \( \alpha = 0.05 \) for each pair-wise comparison.

\subsection{Implementation Details}

All experiments were implemented in Python 3.8, leveraging the following libraries:

\begin{itemize}
    \item \textbf{scikit-learn} \cite{pedregosa2011scikit} was used for data preprocessing, model training, evaluation metrics, PCA, and feature selection.
    \item \textbf{XGBoost} \cite{chen2016xgboost} and \textbf{CatBoost} \cite{dorogush2018catboost} were employed for the respective classifiers.
    \item \textbf{Hugging Face Transformers} \cite{wolf2020transformers} facilitated the generation of contextual embeddings using pre-trained RoBERTa and GPT-2 models.
    \item \textbf{Matplotlib} and \textbf{Seaborn} were utilized for data visualization, including feature importance plots and ROC curves.
\end{itemize}

The Experiments were conducted on a machine equipped with an Intel Core i7 processor, 32 GB RAM, and an NVIDIA GeForce RTX 4050 GPU. GPU acceleration was utilized for embedding generation to expedite processing times. Additionally, all software dependencies were managed using \texttt{conda} environments to ensure reproducibility.\\

We encountered several practical challenges during the experiments, necessitating strategies to address memory management, processing efficiency, and reproducibility, as outlined below: 

\begin{itemize}
    \item \textbf{Memory management} can be challenging when handling large datasets with high-dimensional embeddings. To mitigate memory issues and avoid crashes, we implemented all relevant data-processing steps as batch-wise processing.
    \item The \textbf{processing time} can also be an issue when generating embeddings for large datasets; therefore, in addition to batch processing, we leveraged vectorized operations with libraries like NumPy and Pandas.
    \item For \textbf{reproducibility} purposes and to ensure our results were consistent across different runs, we set fixed random seeds for all libraries involved in stochastic processes. This included setting seeds to a fixed value for NumPy, TensorFlow, PyTorch, and the random module.
\end{itemize}

\section{Performance Evaluation}

This section presents the results of our evaluation of embedding-based feature enrichment on ensemble learning classifiers, specifically Random Forest, XGBoost, and CatBoost, across the eight benchmark datasets presented in Table~\ref{tab:dataset_overview}. Based on a detailed ablation study, we evaluated the effects of combining traditional features with embeddings from pre-trained language models, specifically RoBERTa and GPT-2. All results are presented in this section, with further details in the Appendix, Table~\ref{tab:all-results}. The integration of these contextual embeddings demonstrated significant improvements in predictive performance, particularly in datasets with limited representativeness, i.e., imbalanced datasets or datasets with limited features or samples.

\begin{figure}[hbp]
    \centering
    \makebox[\textwidth]{ 
    \resizebox{1.3\textwidth}{!}{
    \begin{minipage}{0.5\textwidth}
        \centering
        \includegraphics[width=\linewidth, height=6.5cm]{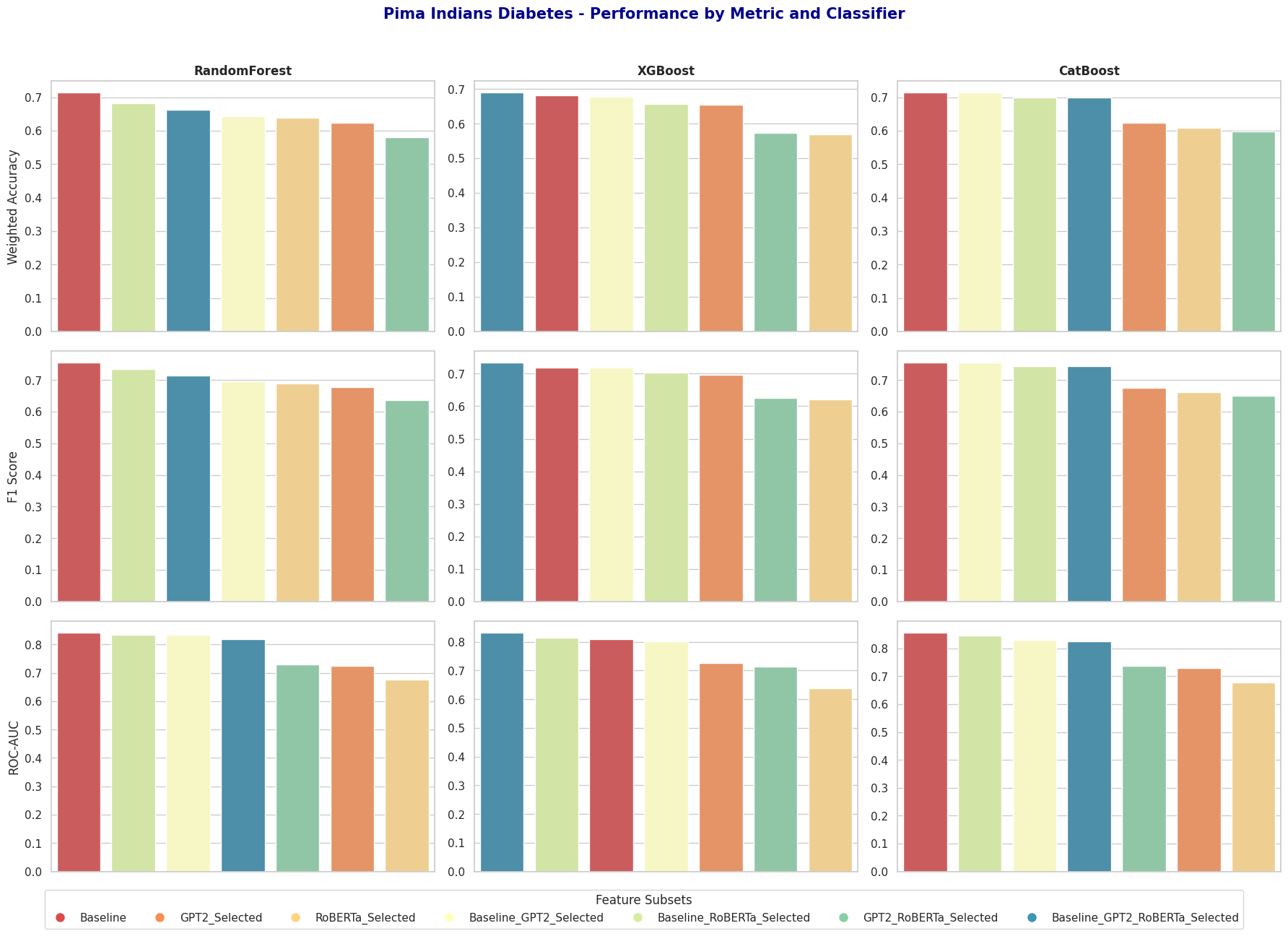}
    \end{minipage}\hfill
    \begin{minipage}{0.5\textwidth}
        \centering
        \includegraphics[width=\linewidth, height=6.5cm]{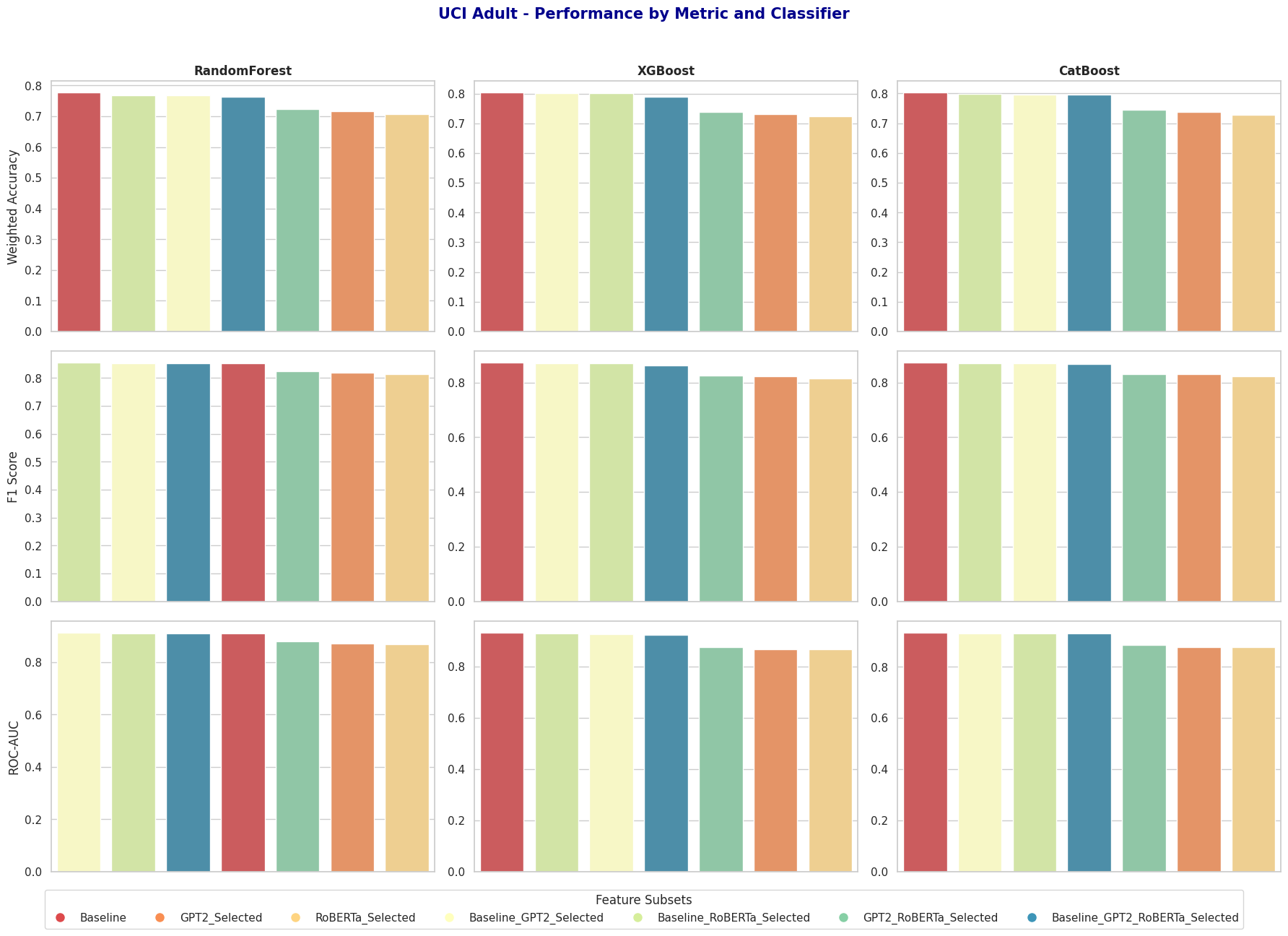}
    \end{minipage}
    }
    }
    \makebox[\textwidth]{ 
    \resizebox{1.3\textwidth}{!}{
    \begin{minipage}{0.5\textwidth}
        \centering
        \includegraphics[width=\linewidth, height=6.5cm]{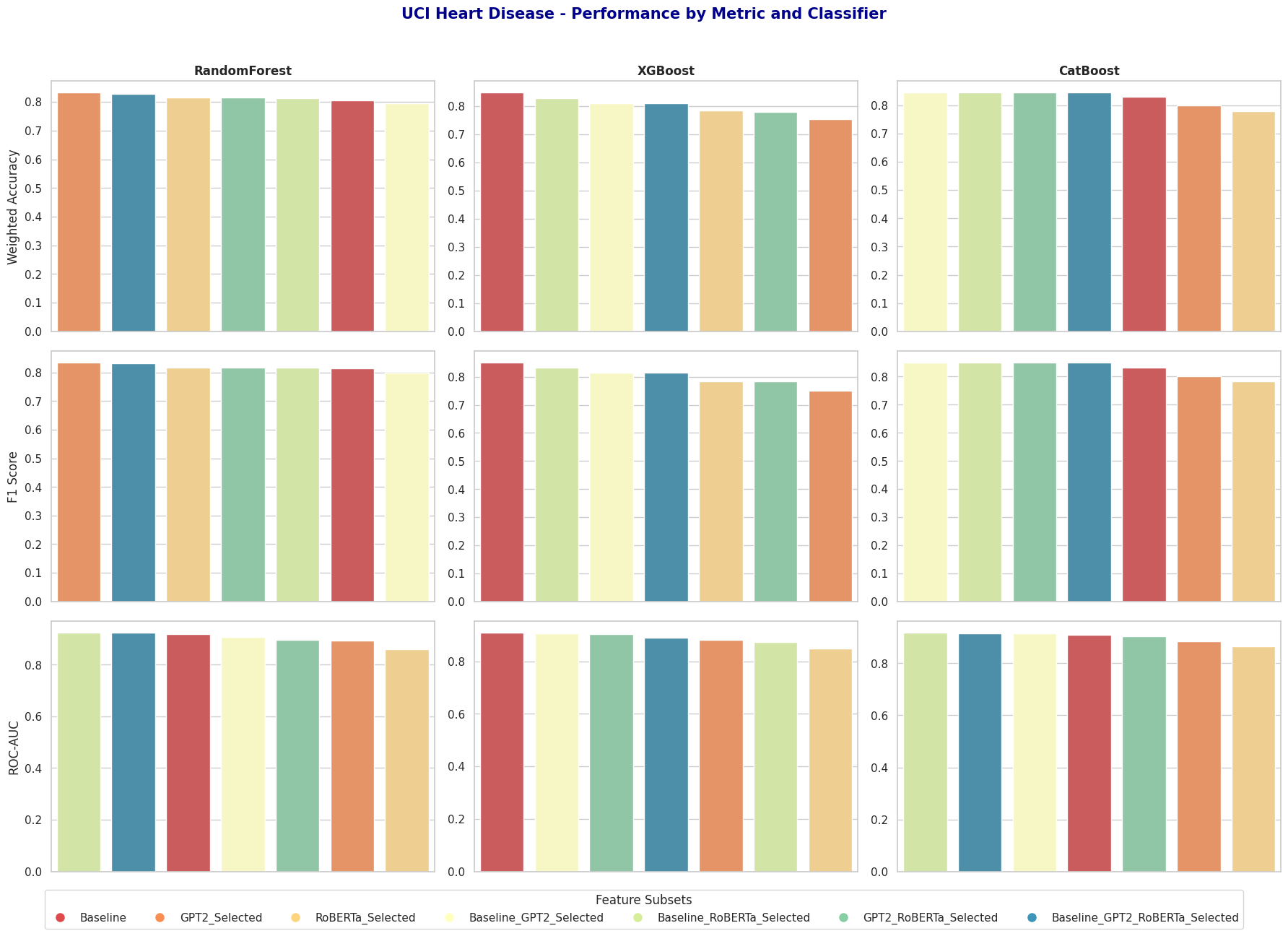}
    \end{minipage}\hfill
    \begin{minipage}{0.5\textwidth}
        \centering
        \includegraphics[width=\linewidth, height=6.5cm]{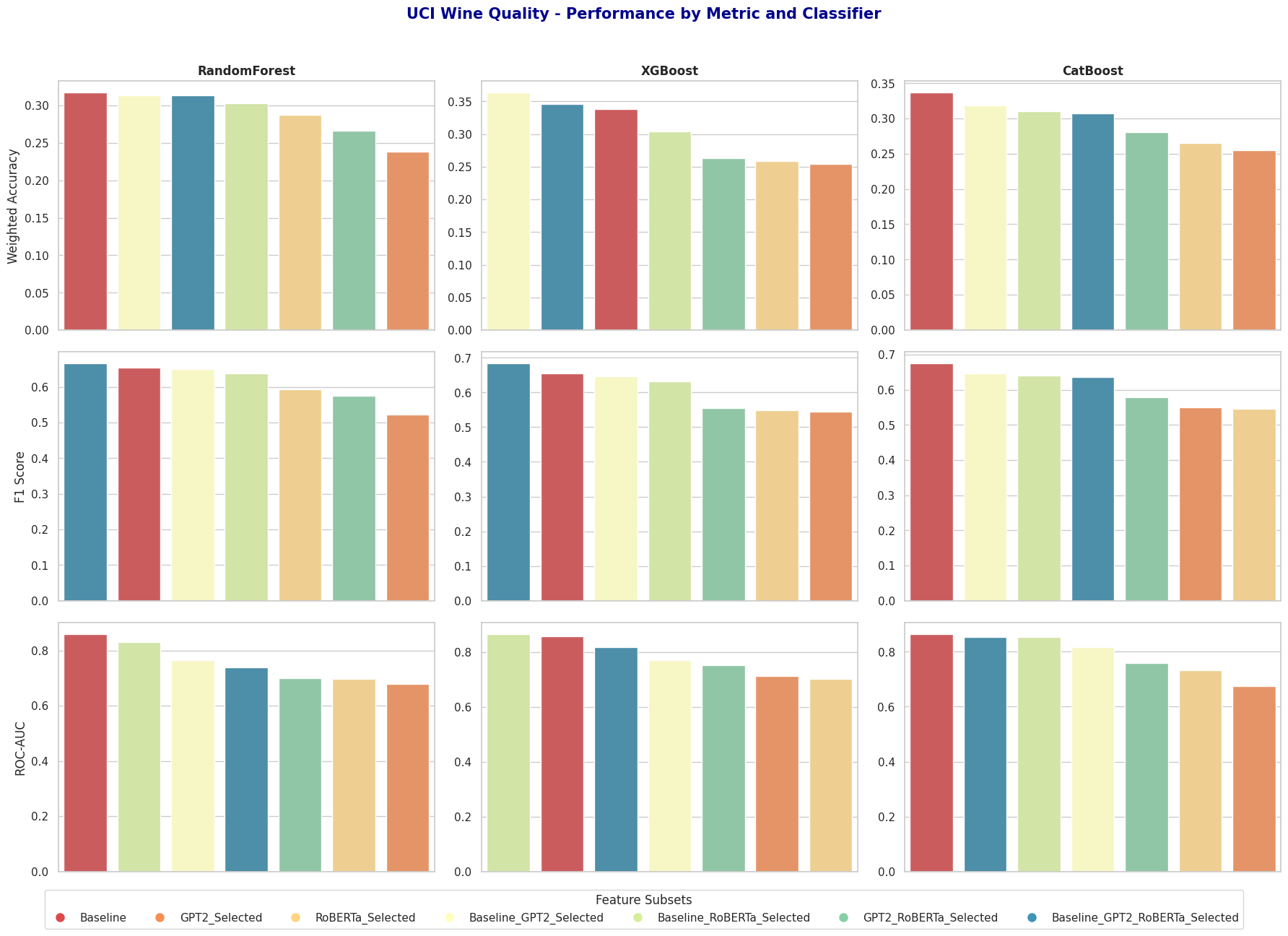}
    \end{minipage}
    }
    }
    \caption{Performance metrics (Weighted/Balanced Accuracy, F1 Score, and ROC-AUC) for different feature subsets on multiple datasets (\texttt{Pima Indians Diabetes}, \texttt{UCI Adult}, \texttt{UCI Heart Disease}, and \texttt{UCI Wine Quality}) using Random Forest, XGBoost, and CatBoost classifiers. Each plot shows how different feature subsets influence classifier performance, with results sorted by feature subset to facilitate comparisons.}
\label{fig:combined_feature_performance_plots}
\end{figure}

\subsection{Overall Performance Trends Across Models and Feature Subsets}
\begin{figure}[h!]
\centering
\includegraphics[width=0.9\linewidth]{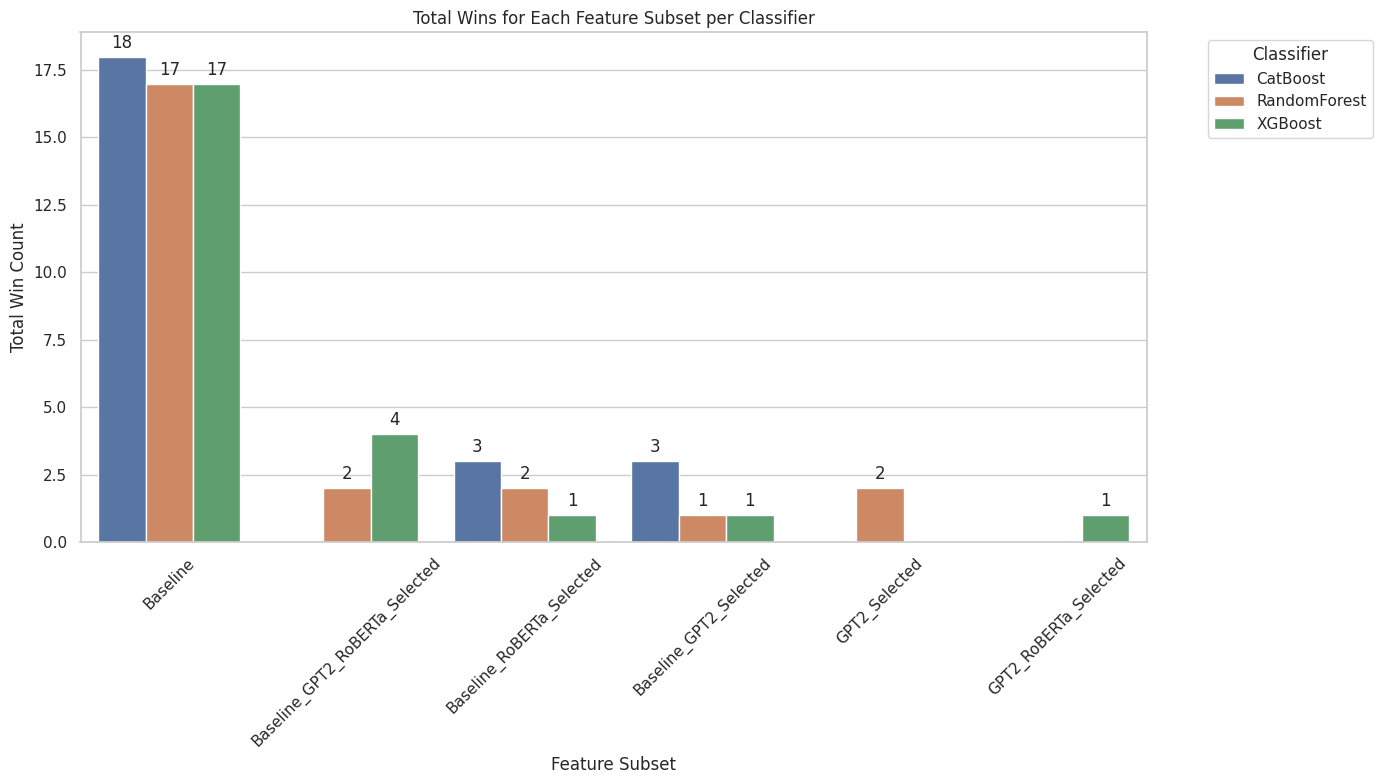}
\caption{Total wins for each feature subset per classifier.}
    \label{fig:win_counts}
\end{figure}

\paragraph{Performance Trends.} As shown in Figure~\ref{fig:combined_feature_performance_plots}, in many datasets, feature subsets that incorporate embeddings show clear improvements in both Weighted/Balanced Accuracy,  F1, and ROC-AUC Score over the baseline features alone. For example, in the \texttt{UCI Adult}, \texttt{Pima Indian Diabetes}, \texttt{Titanic} and \texttt{Heart Disease datasets}, feature subsets enhanced with GPT-2 and RoBERTa embeddings yield significant gains, especially in combinations with more advanced ensemble classifiers, such as XGBoost and CatBoost. These results suggest that the embedding-based enrichments capture essential information that improve classification. Moreover, XGBoost and CatBoost benefit from the enriched features more often than the Random Forest classifier. This can also be seen in Figure~\ref{fig:win_counts}, which shows the absolute counts of wins per classifier, per feature subset.

\paragraph{Key Takeaways.} A key insight from these results is that for datasets with limited representativeness -- such as those found in medical applications -- the computational overhead in embedding-based feature enrichment can be well justified by the performance gains achieved. Embeddings add valuable semantic context, compensating for the scarcity of features or class diversity, and thus can enhance model performance. The gains from the embedding-derived features can be further amplified with more advanced ensemble classifiers such as XGBoost or CatBoost. 

Conversely, for datasets with robust representation across classes and features, the benefits of embedding enrichment tend to be marginal at best, as these datasets already offer sufficient information for the model to learn effectively.

\subsection{Statistical Significance of Performance Improvements}

To assess whether the integration of contextual embeddings (i.e., GPT-2 and RoBERTa) yields significant improvements in model performance compared to baseline features, we conducted a series of Bonferroni-corrected paired t-tests across all classifiers and datasets used in this study. These t-tests compared different feature subsets to determine if embedding-based enrichments had a statistically significant effect (\( p < 0.05 \)) on metrics such as accuracy, balanced accuracy, F1 score, and ROC-AUC. \\

\begin{figure}[h!]
    \centering
    \begin{minipage}{\textwidth}
        \centering
    \includegraphics[width=\linewidth]{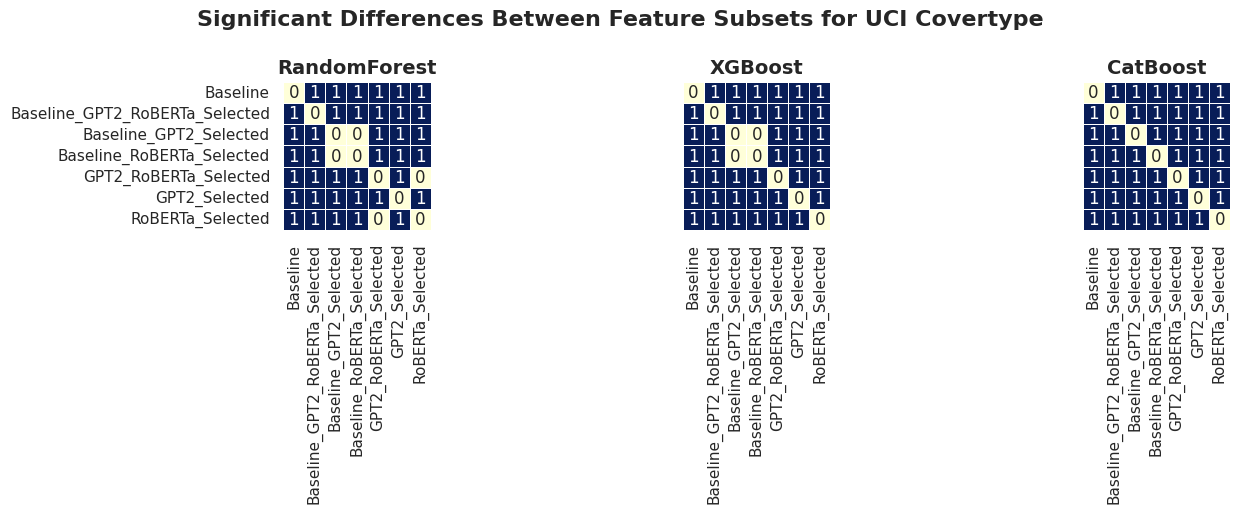}
    \end{minipage}
    \vspace{-10pt} 
    \begin{minipage}{\textwidth}
        \centering
   \includegraphics[width=\linewidth]{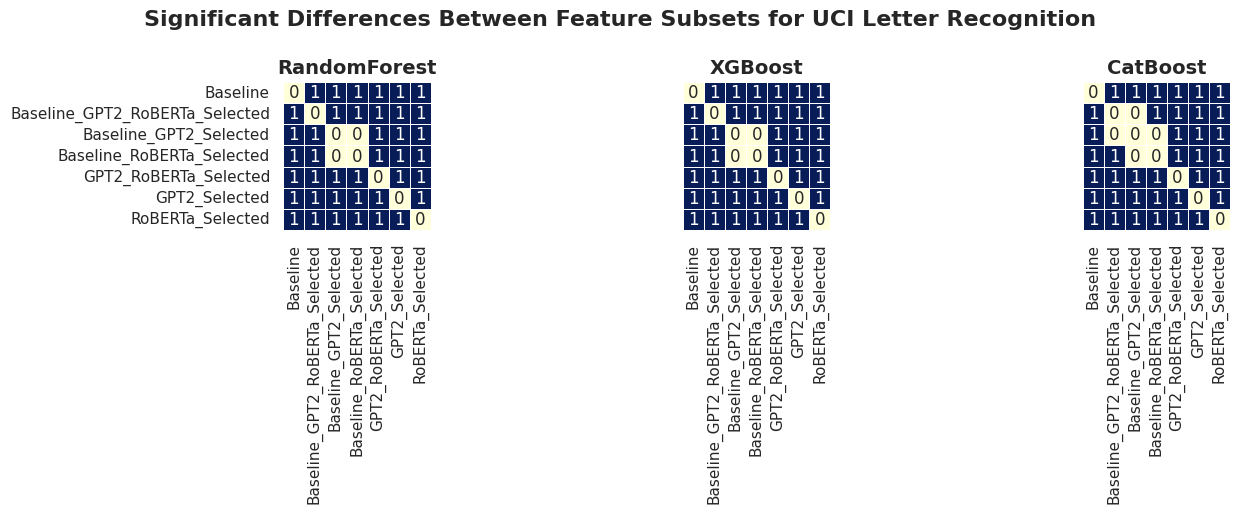}
    \end{minipage}
    \vspace{-10pt} 
    \begin{minipage}{\textwidth}
        \centering
    \includegraphics[width=\linewidth]{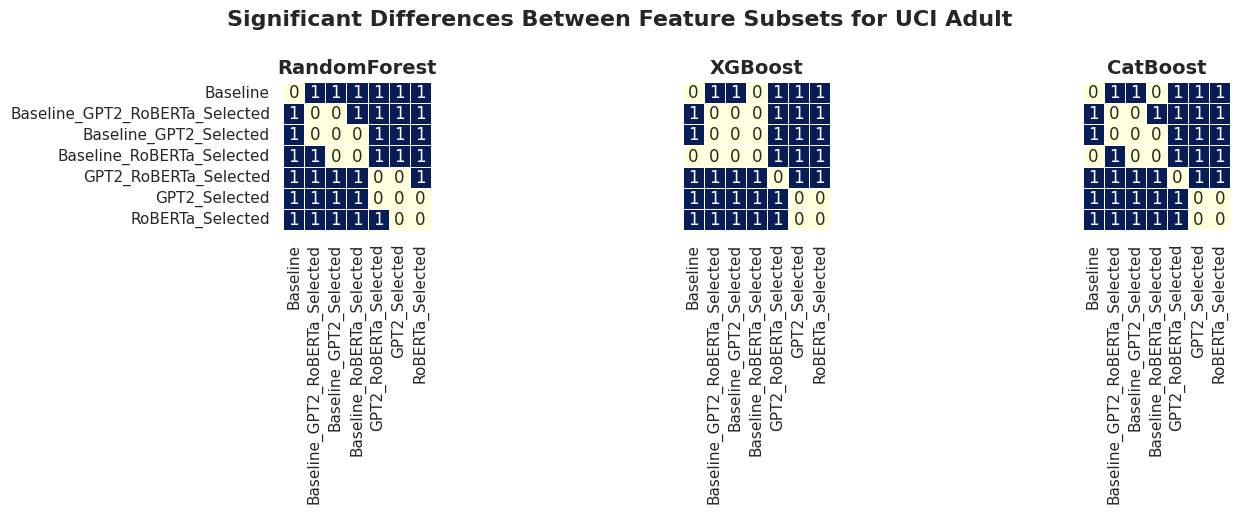}
    \end{minipage}
    \caption{Significant differences between feature subsets across the datasets \texttt{UCI Covertype}, \texttt{UCI Letter Recognition}, and \texttt{UCI Adult}. Each subplot shows pairwise statistical significance tests, where dark cells (value 1) indicate statistically significant differences between feature subsets. The results demonstrate how embeddings (e.g., GPT2, RoBERTa) impact model performance differently across datasets and classifiers.}
\label{fig:combined_significance_plots_1}
\end{figure}

\begin{figure}[h!]
    \centering
    \begin{minipage}{\textwidth}
        \centering
    \includegraphics[width=\linewidth]{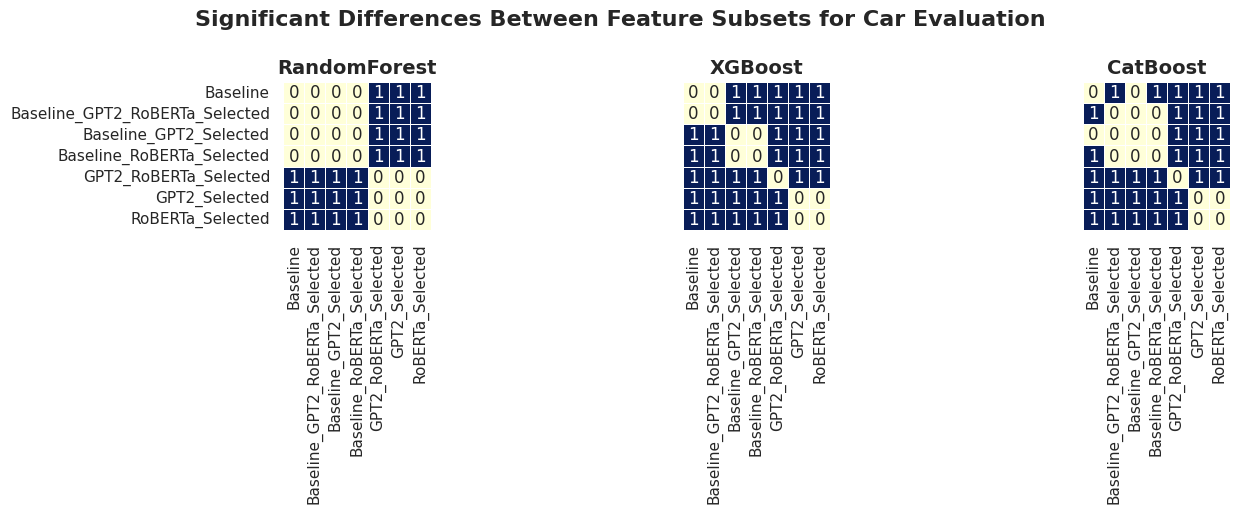}
    \end{minipage}
    \vspace{-10pt} 
    \begin{minipage}{\textwidth}
        \centering
   \includegraphics[width=\linewidth]{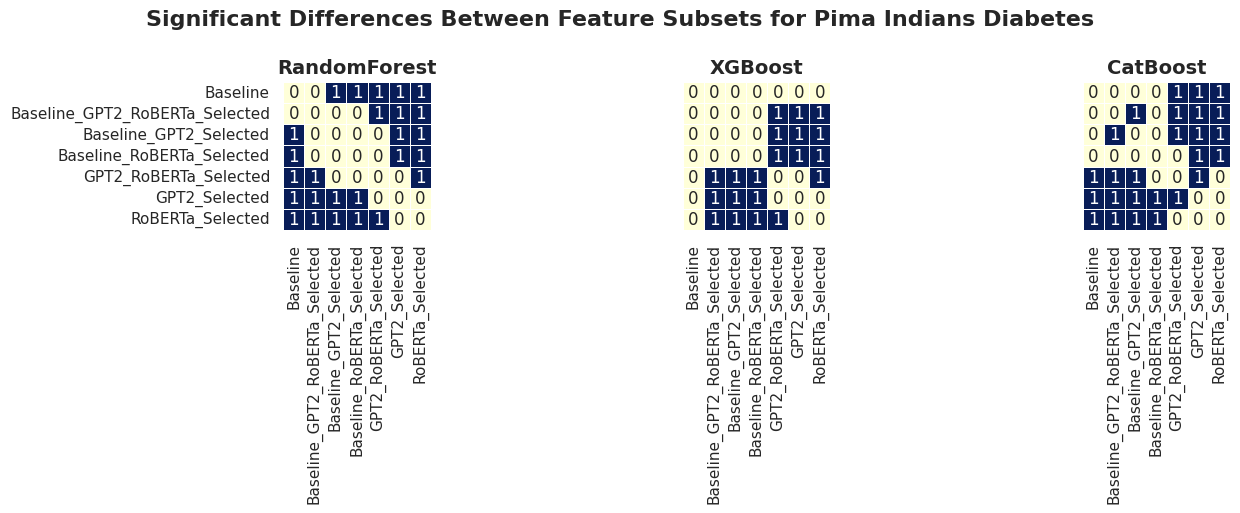}
    \end{minipage}
    \vspace{-10pt} 
    \begin{minipage}{\textwidth}
        \centering
    \includegraphics[width=\linewidth]{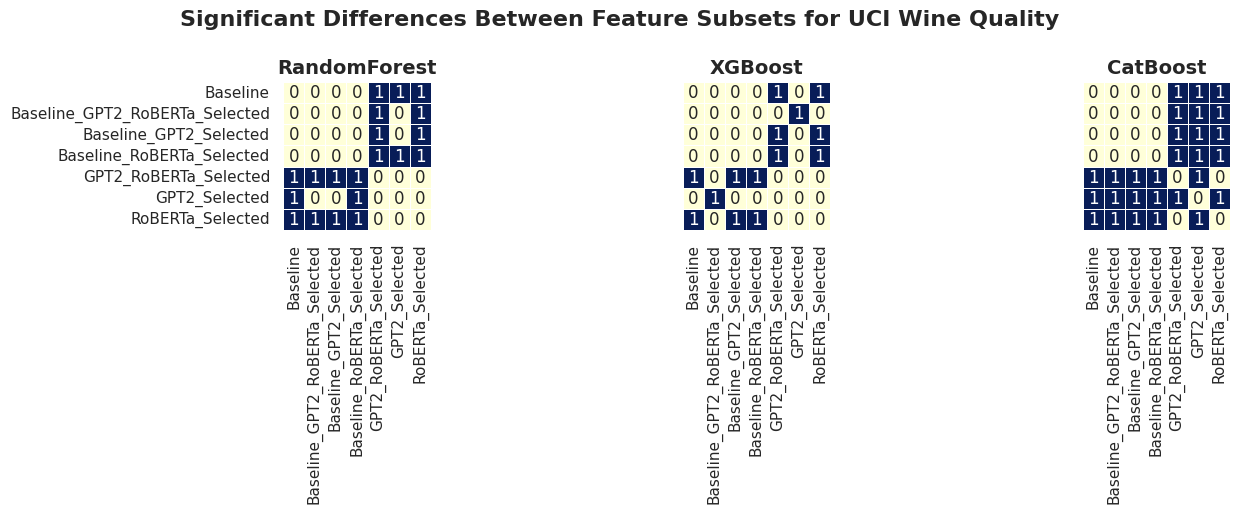}
    \end{minipage}
    \caption{Significant differences between feature subsets across the datasets \texttt{Car Evaluation}, \texttt{Pima Indians Diabetes}, and \texttt{UCI Wine Quality}. Each subplot shows pairwise statistical significance tests, where dark cells (1) indicate statistically significant differences between feature subsets. The results demonstrate how embeddings (e.g., GPT2, RoBERTa) impact model performance differently across datasets and classifiers.}
\label{fig:combined_significance_plots2}
\end{figure}

\begin{figure}[h!]
    \centering
    \begin{minipage}{\textwidth}
        \centering
    \includegraphics[width=\linewidth]{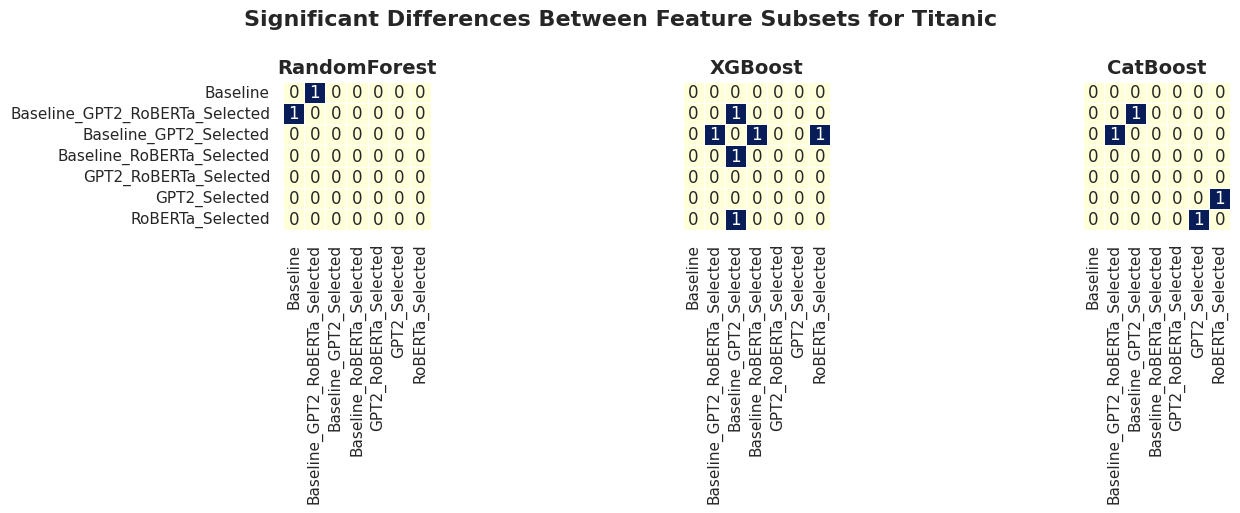}
    \end{minipage}
    \vspace{5pt} 
    \begin{minipage}{\textwidth}
        \centering
    \includegraphics[width=\linewidth]{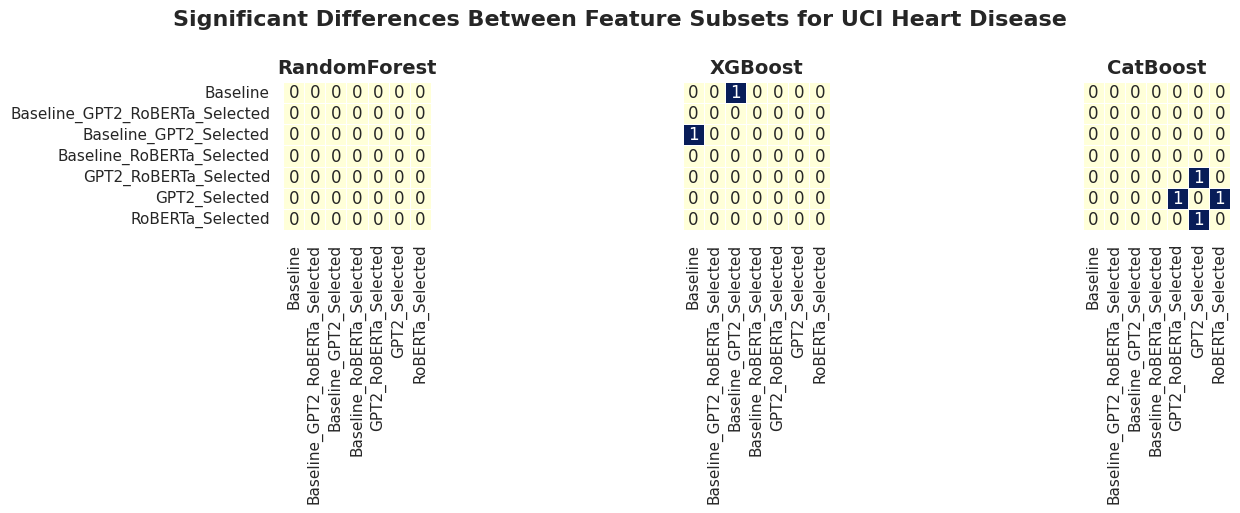}
    \end{minipage}
    \caption{Significant differences between feature subsets across the datasets \texttt{Titanic} and \texttt{UCI Heart Disease}. Each subplot shows pairwise statistical significance tests, where dark cells (1) indicate statistically significant differences between feature subsets. The results demonstrate how embeddings (e.g., GPT2, RoBERTa) impact model performance differently across datasets and classifiers.}
\label{fig:combined_significance_plots3}
\end{figure}

Figures~\ref{fig:combined_significance_plots_1}, \ref{fig:combined_significance_plots2}, and \ref{fig:combined_significance_plots3} from the Appendix illustrate pairwise statistical significance analyses among feature subsets across the eight benchmark datasets. Each subplot shows a matrix of pairwise comparisons, where the value of 1 in dark-colored cells indicates that the differences between feature subsets are statistically significant (\( p < 0.05 \), after Bonferroni correction).\\

 In comparison to subsets involving only baseline features, we often found significant performance enhancements in feature subsets enriched with GPT-2 and RoBERTa embeddings, mostly in combination with XGBoost and CatBoost.  However, the performance gains were varied, which suggests that while contextual embeddings may offer benefits for the predictive performance, the extent of those benefits may depend on dataset characteristics such as class representation and balance, dataset complexity, and feature diversity. For the \texttt{UCI Letter Recognition} and \texttt{Covertype} datasets, we did not observe any gains from embedding-derived features, no matter which classifier or which enriched feature subset was used. We hypothesize that these two datasets already possess highly informative and well-defined feature sets that effectively capture the underlying patterns needed for classification. When datasets are structured and have robust representations, embeddings may not add significant value, as the original features alone suffice for accurate predictions.

\subsection{Feature Importance Analysis}

\begin{figure}[h!]
    \centering
    \makebox[\textwidth]{ 
    \resizebox{1.3\textwidth}{!}{
    \begin{minipage}{0.33\textwidth}
        \centering
        \includegraphics[width=\linewidth]{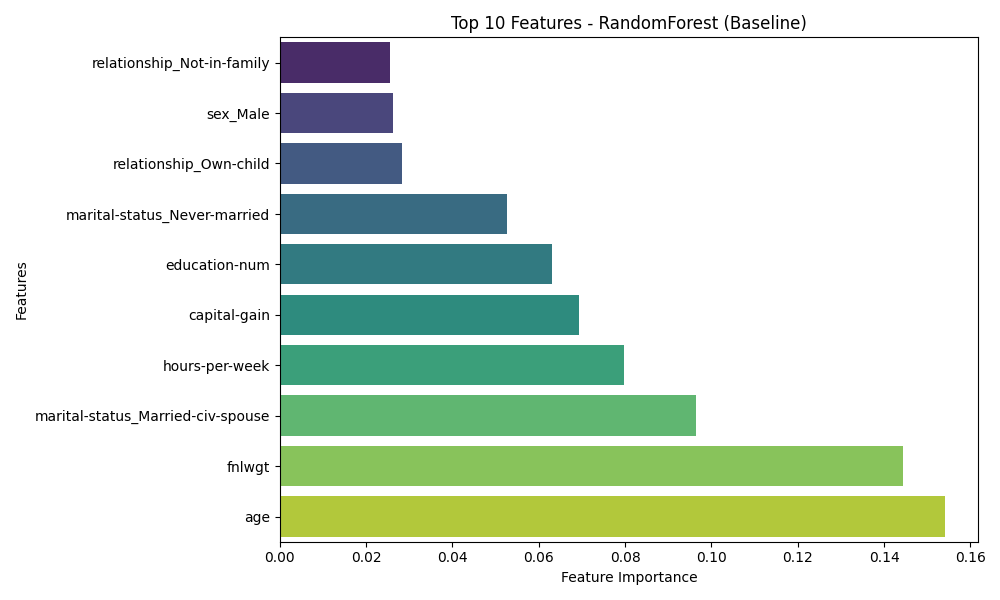}
    \end{minipage}\hfill
    \begin{minipage}{0.33\textwidth}
        \centering
        \includegraphics[width=\linewidth]{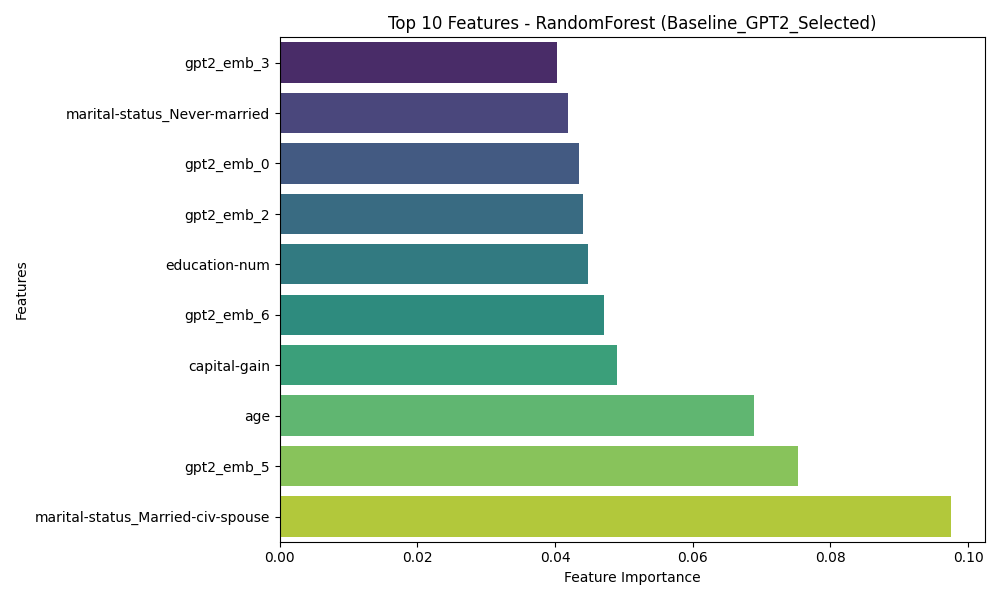}
    \end{minipage}\hfill
    \begin{minipage}{0.33\textwidth}
        \centering
        \includegraphics[width=\linewidth]{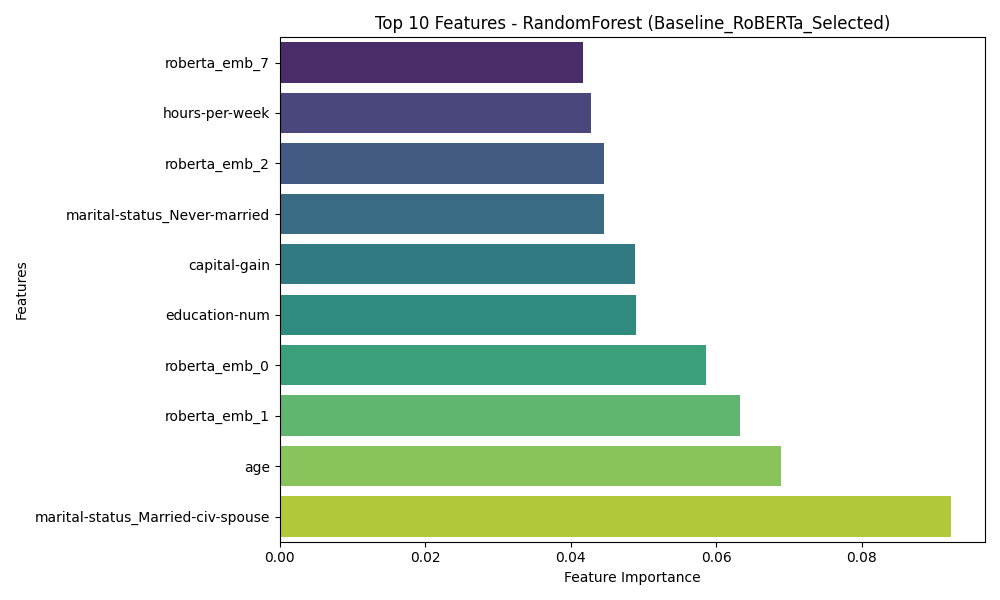}
    \end{minipage}
    }
    }
    \caption{Random Forest top-10 features for the three best performing feature subsets on the \texttt{UCI Adult dataset}.}
    \label{fig:Top-10_features_three_best_performing_feature_subsets_Adult}
\end{figure}

\begin{figure}[h!]
    \centering
    \makebox[\textwidth]{ 
    \resizebox{1.3\textwidth}{!}{
    \begin{minipage}{0.33\textwidth}
        \centering
        \includegraphics[width=\linewidth]{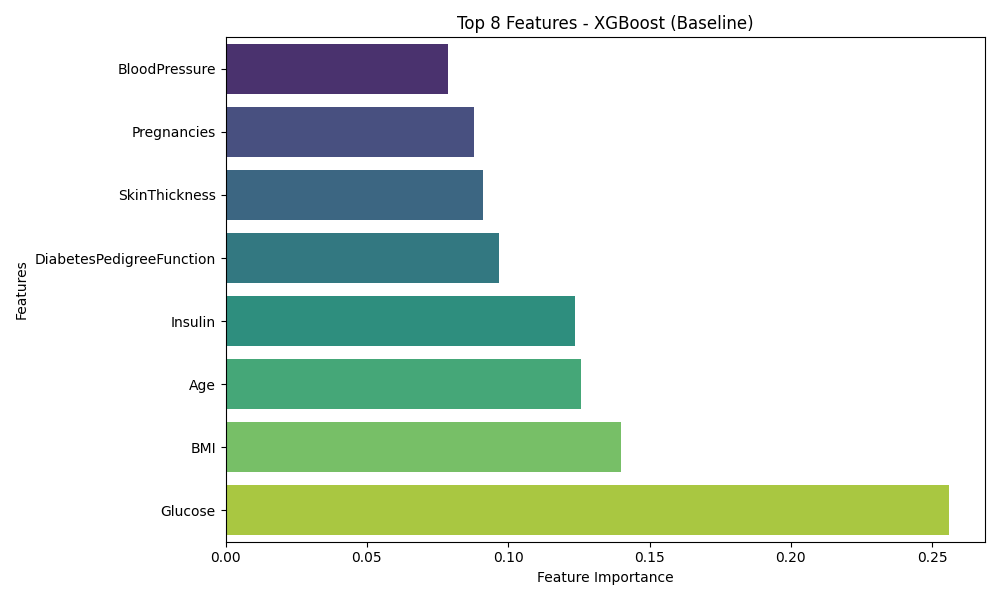}
    \end{minipage}\hfill
    \begin{minipage}{0.33\textwidth}
        \centering
        \includegraphics[width=\linewidth]{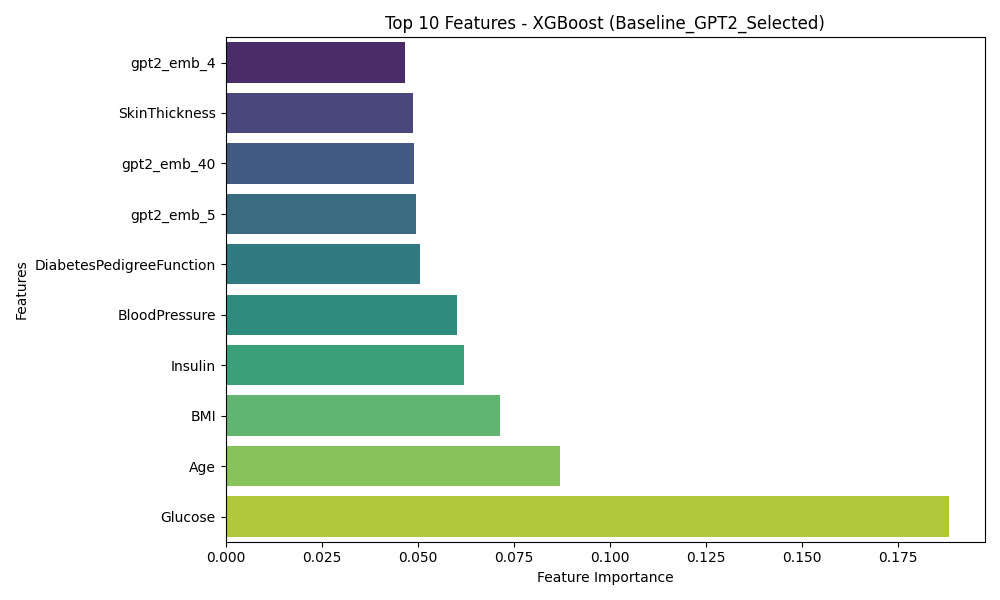}
    \end{minipage}\hfill
    \begin{minipage}{0.33\textwidth}
        \centering
        \includegraphics[width=\linewidth]{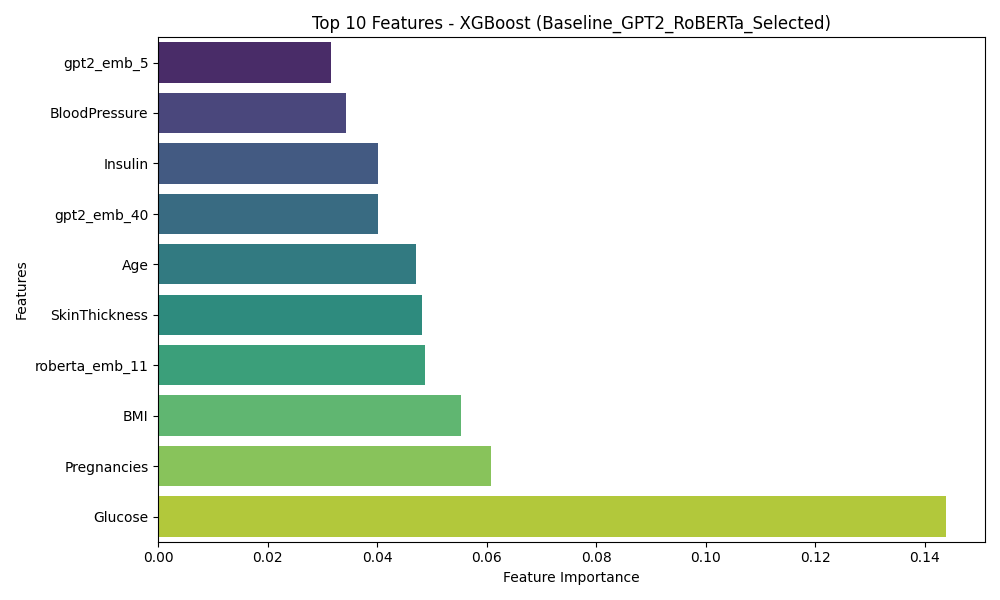}
    \end{minipage}
    }
    }
    \caption{XGBoost top-10 features for the three best-performing feature subsets on the \texttt{UCI Pima Indian Diabetes} dataset.}
    \label{fig:Top-10_features_three_best_performing_feature_subsets_Indian_Diabetes}
\end{figure}

\begin{figure}[h!]
    \centering
    \makebox[\textwidth]{ 
    \resizebox{1.25\textwidth}{!}{
    \begin{minipage}{0.33\textwidth}
        \centering
        \includegraphics[width=\linewidth]{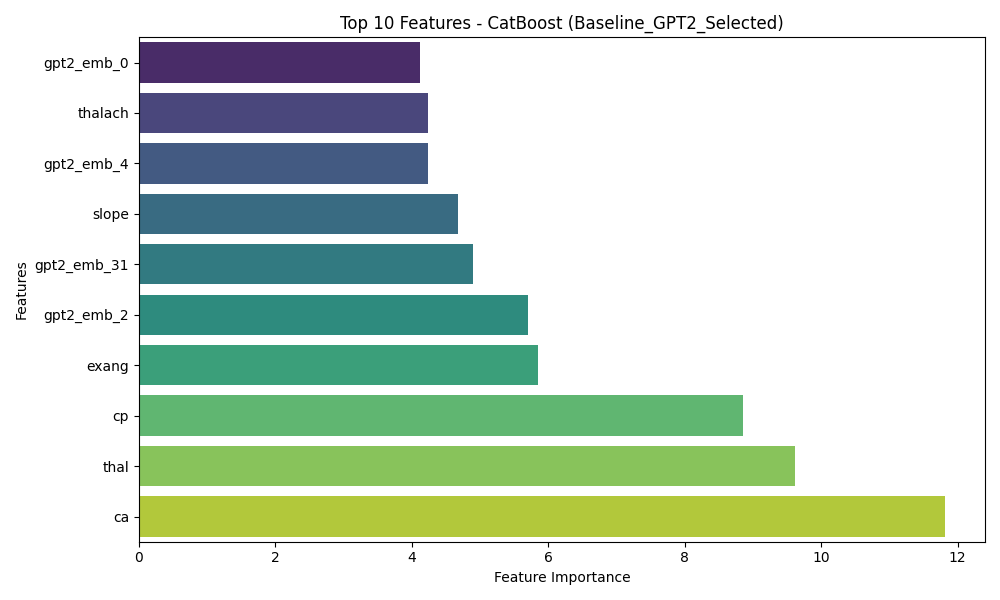}
    \end{minipage}\hfill
    \begin{minipage}{0.33\textwidth}
        \centering
        \includegraphics[width=\linewidth]{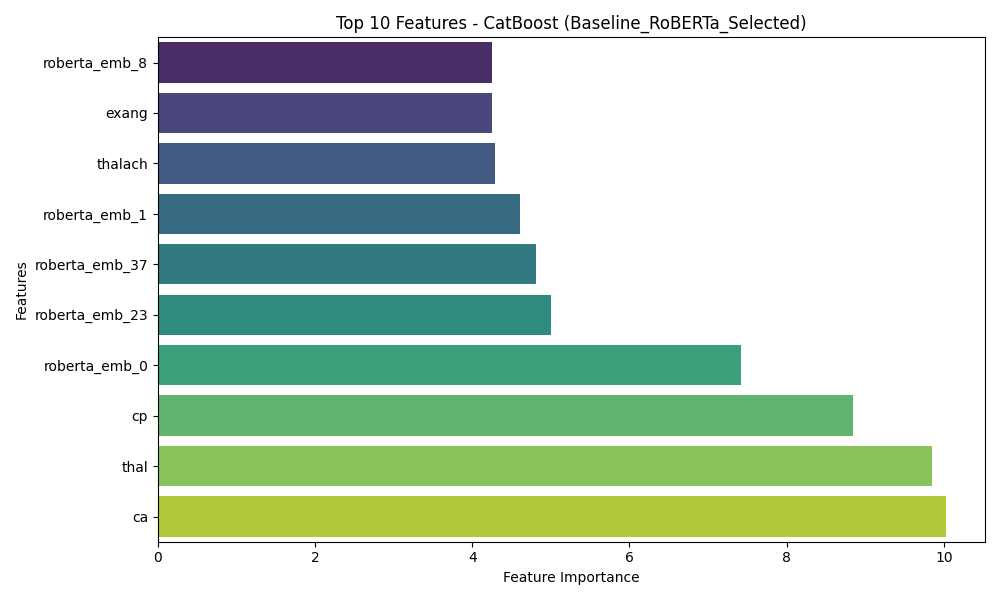}
    \end{minipage}\hfill
    \begin{minipage}{0.33\textwidth}
        \centering
        \includegraphics[width=\linewidth]{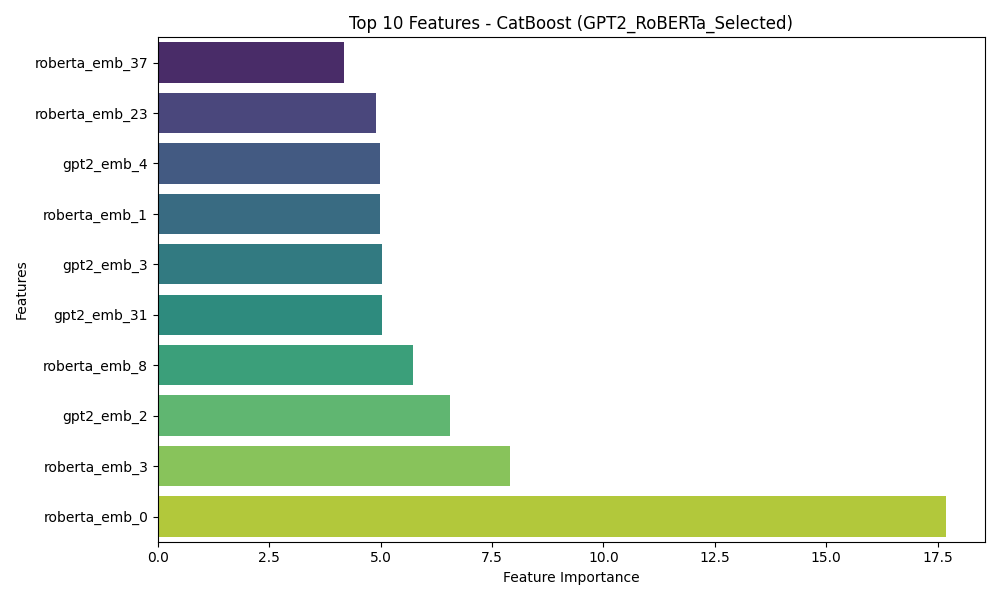}
    \end{minipage}
    }
    }
    \caption{CatBoost top-10 features for the three best performing feature subsets on the \texttt{UCI Heart Disease} dataset.}
    \label{fig:Top-10_features_three_best_performing_feature_subsets_UCI_Heart}
\end{figure}

In order to understand the relative impact of different feature subsets on model predictions, we conducted an extensive feature importance analysis, with importance scores derived from the Random Forest, XGBoost, and CatBoost classifiers. Figures~\ref{fig:Top-10_features_three_best_performing_feature_subsets_Adult},~\ref{fig:Top-10_features_three_best_performing_feature_subsets_Indian_Diabetes}, and~\ref{fig:Top-10_features_three_best_performing_feature_subsets_UCI_Heart} present the top 10 features for the three best-performing feature subsets across three selected datasets, namely the \texttt{UCI Adult}, \texttt{Pima Indians Diabetes}, and \texttt{Heart Disease}.

Across classifiers and datasets, embedding-derived features, such as those from GPT-2 and RoBERTa, frequently rank among the most important, which illustrates their significant contribution to classification outcomes. For example, in Figure\ref{fig:Top-10_features_three_best_performing_feature_subsets_Adult}, GPT-2 and RoBERTa features appear prominently among the top-10-ranked features for the \texttt{UCI Adult} dataset, indicating that these embeddings effectively capture essential information that complements the original tabular features. 

In the case of the \texttt{Pima Indians Diabetes} dataset, as shown in Figure~\ref{fig:Top-10_features_three_best_performing_feature_subsets_Indian_Diabetes}, the XGBoost classifier reveals a similar trend where embedding-based features contribute to the model’s output beyond the capacity of baseline features alone. This underscores the impact of contextual embeddings in enriched feature subsets.

For the \texttt{UCI Heart Disease} dataset, Figure~\ref{fig:Top-10_features_three_best_performing_feature_subsets_UCI_Heart} highlights the impact of embedding-enriched features using the CatBoost classifier. For this dataset, the combination of baseline features and embeddings provides a richer representation. However, interestingly, the CatBoost classifier performed best on the combination features derived from GPT-2 and RoBERTa embeddings alone. This suggests that the embeddings effectively capture medically relevant information that the original features might miss.

As showcased by these examples, embeddings generated from pretrained LLMs may provide a complementary view by incorporating additional semantic information and lead to improved decision-making by ensemble classifiers. The consistent presence of these features among the top ranks suggests that their integration is deemed important by the underlying model and might help enhance representation diversity.

We believe that such hybrid approaches have the potential to advance feature engineering practices, especially in ensemble learning frameworks, by leveraging the strengths of both feature types to improve predictive performance and generalizability.

\subsection{Visualization of Embedding Distributions}

\begin{figure}[h!]
    \centering
    \makebox[\textwidth]{ 
    \resizebox{1.15\textwidth}{!}{
    \begin{minipage}{0.5\textwidth}
        \centering
        \includegraphics[width=\linewidth]{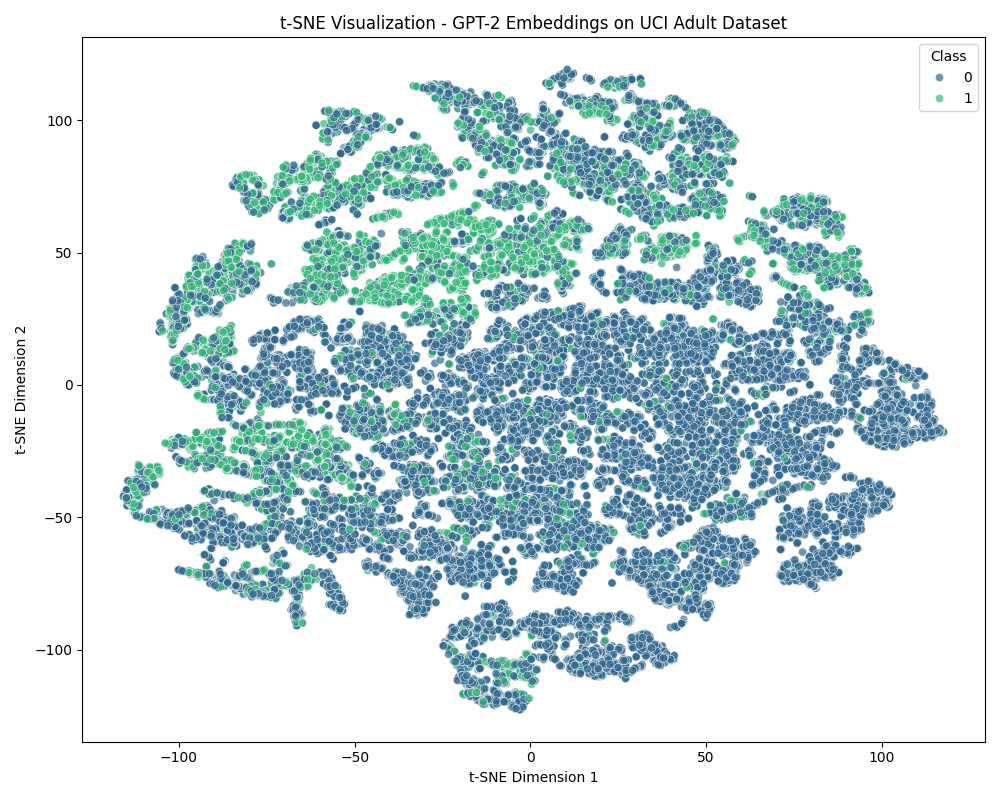}
    \end{minipage}\hfill
    \begin{minipage}{0.5\textwidth}
        \centering
        \includegraphics[width=\linewidth]{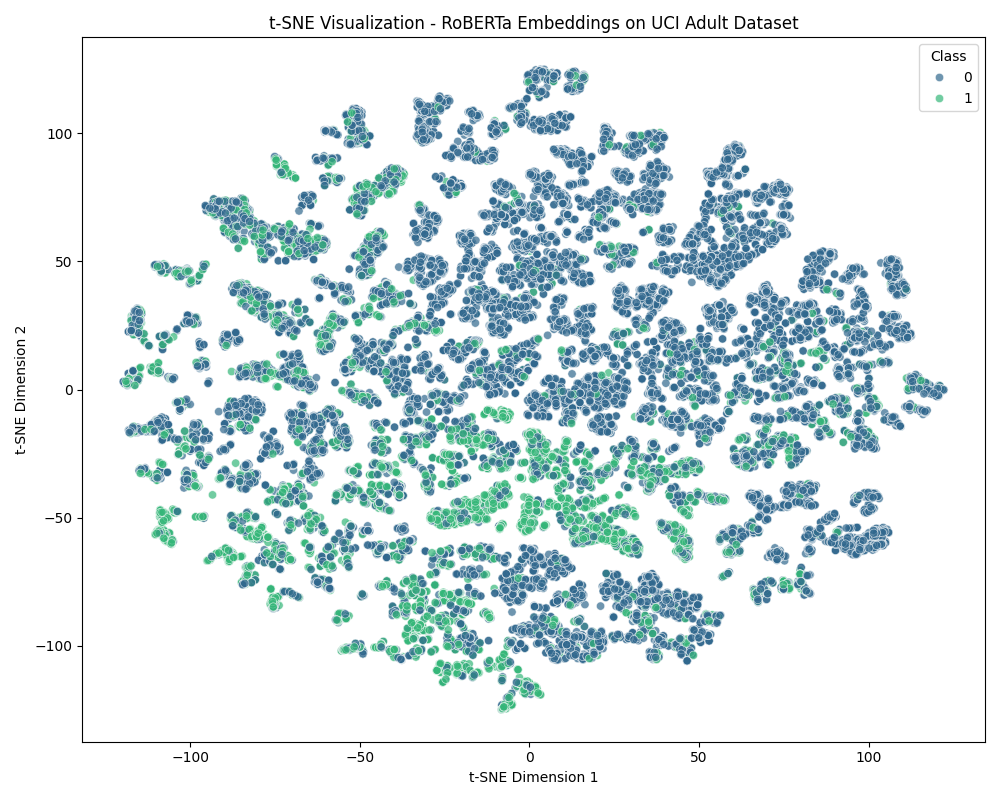}
    \end{minipage}
    }
    }
    \caption{t-SNE applied to the selected GPT-2 and RoBERTa embeddings from the UCI Adult dataset for 2D visualization.}
\label{fig:tSNE_Embeddigns_Adult}
\end{figure}

\begin{figure}[h!]
    \centering
    \makebox[\textwidth]{ 
    \resizebox{1.15\textwidth}{!}{
    \begin{minipage}{0.5\textwidth}
        \centering
        \includegraphics[width=\linewidth]{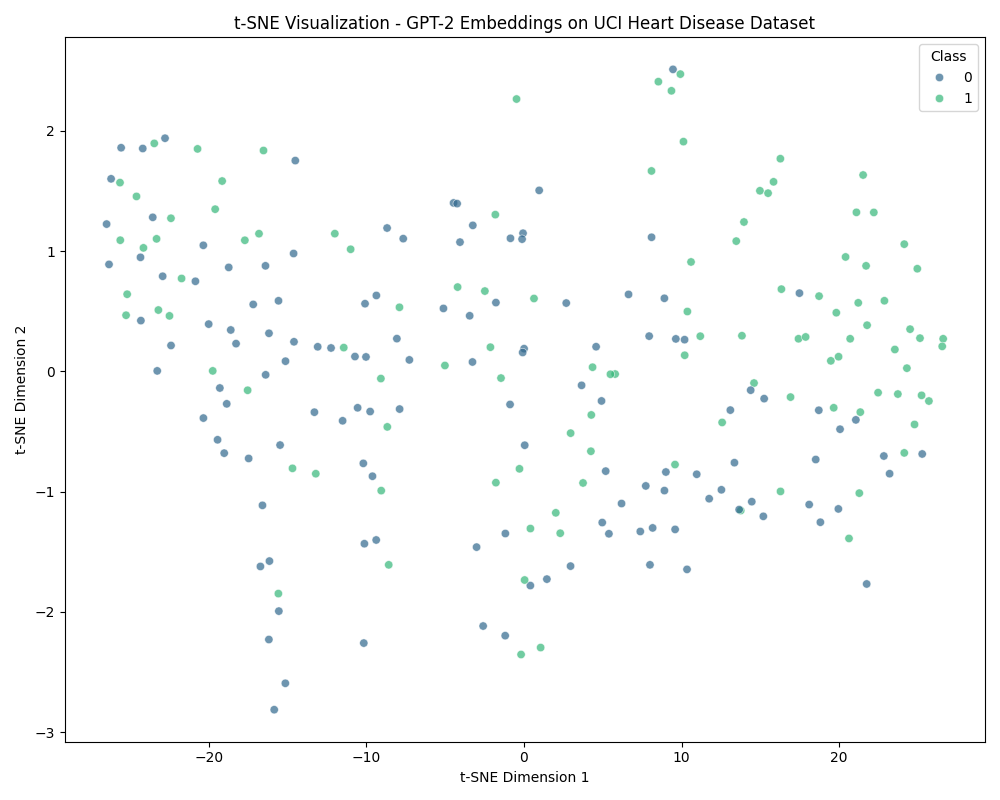}
    \end{minipage}\hfill
    \begin{minipage}{0.5\textwidth}
        \centering
        \includegraphics[width=\linewidth]{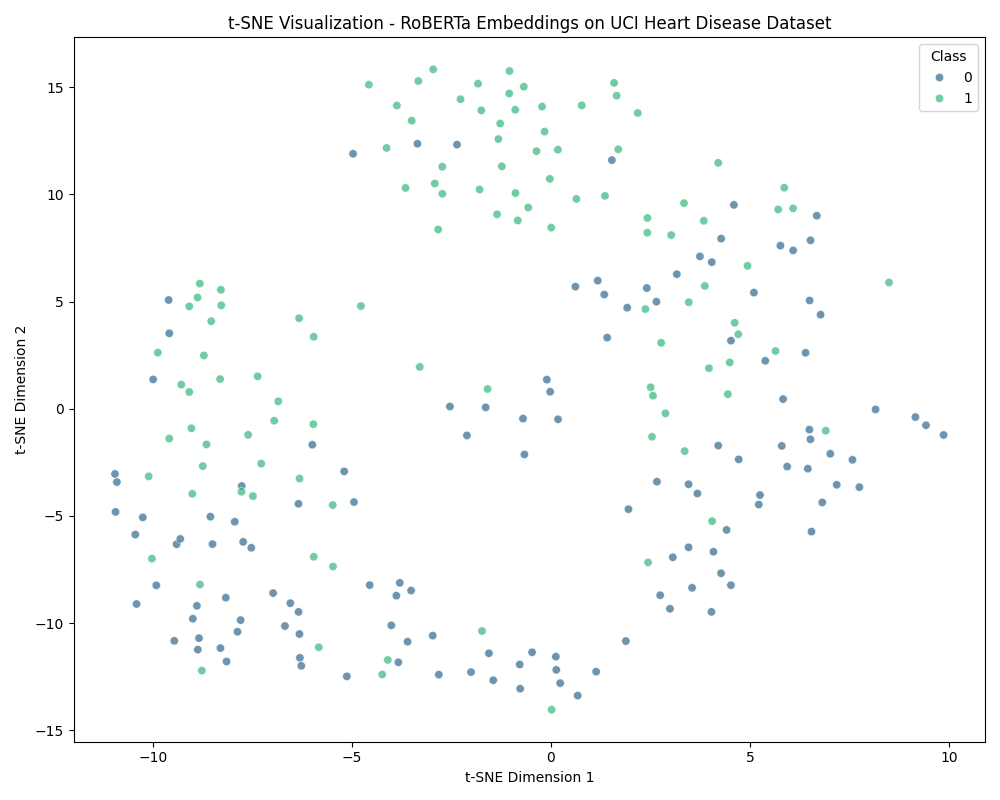}
    \end{minipage}
    }
    }
    \caption{t-SNE applied to the selected GPT-2 and RoBERTa embeddings from the UCI Heart Disease dataset for 2D visualization.}
    \label{fig:tSNE_Embeddigns_Heart}
\end{figure}

To further illustrate the discriminative power of the generated embeddings, we employed t-distributed Stochastic Neighbor Embedding (t-SNE)~\cite{van2008visualizing} to project high-dimensional embedding vectors into a 2D space. Figures~\ref{fig:tSNE_Embeddigns_Adult} and~\ref{fig:tSNE_Embeddigns_Heart} depict the clustering of RoBERTa and GPT-2 embeddings with respect to target classes in the \texttt{UCI Adult} and \texttt{UCI Heart Disease} datasets, respectively.


These visualizations reveal clustering patterns corresponding to different target classes, affirming that the embeddings encapsulate discriminative information conducive to accurate classification.

\section{Discussion}

\subsection{Privacy-Preserving Potential of Embeddings}

\paragraph{Privacy Advantages of Embedding-Based Representations.} Embedding-based features present an effective approach to privacy-preserving machine learning, particularly beneficial in domains like healthcare. By abstracting raw data into high-dimensional embeddings, sensitive details are veiled, reducing the risk of data exposure. Embedding-only subsets, such as GPT2\_Selected and RoBERTa\_Selected, often perform well, slightly worse than raw-feature-based models. They align well with privacy-by-design principles and regulatory standards like GDPR and allow for acceptable predictive performance without direct reliance on sensitive features, offering both compliance and robust data protection.

\paragraph{Transferability and Model Privacy.} Leveraging embeddings from pre-trained models like GPT-2 and RoBERTa enhances model flexibility and transferability. These embeddings encapsulate generalizable patterns that protect specific dataset characteristics, enabling safe, effective deployment across applications. By allowing organizations to reuse these representations without exposing unique dataset details, embedding-based enrichment meets both the privacy and versatility needs of modern machine learning workflows.

\paragraph{Regulatory Alignment and Data Minimization.} Embedding-based feature enrichment aligns well with data minimization principles by reducing the model's dependency on raw, sensitive data. This approach lowers the need for ongoing access to personal information, supporting compliance with privacy regulations without sacrificing model quality. Embedding transformations, which enhance both performance and data protection, represent a new paradigm where privacy preservation complements predictive accuracy, setting a high standard for ethical AI.

\subsection{Limitations}

The proposed method faces several limitations. First, embeddings from pretrained LLMs like GPT-2 and RoBERTa generate computational overhead, posing scalability challenges for large datasets or real-time applications. The use of 50 PCA components and selecting the top 10 features were based on preliminary experiments; optimal values could vary, thus necessitating further exploration for better efficiency and effectiveness.

Moreover, embeddings, even after feature selection, remain less interpretable compared to raw features, particularly in sensitive, high-stakes applications. Developing methods to enhance interpretability is crucial. Embeddings can also amplify biases in noisy or inconsistently encoded datasets. For low-diversity datasets, there exists the risk of embedding degeneration, limiting the ability to distinguish between classes ore relevant groups in an unbiased way.

Embeddings may also add unnecessary complexity in structured tabular data with minimal contextual information, where their informative value is limited, as seen in some of the datasets used in this study. Future work could further investigate the expected use-case and dataset-specific value of embedding-derived features.

\subsection{Future Research Directions}

In alignment with the above limitations, we see high potential for research that focuses on the following areas: The development of use-case-specific embedding strategies for structured data, boosting the semantic relevance of embedding-derived features, could be highly impactful. Additionally, adaptive feature-weighting and regularization methods could be integrated to enhance robustness and prevent embedding degeneration in complex datasets or datasets with limited representativeness of features or classes.

Improving the interpretability of embeddings is a promising avenue for future research, particularly in privacy-sensitive, high-stakes applications, by developing transparency and privacy-preserving frameworks for structured data. To reduce computational overhead, embedding alternatives could be explored alongside automated, early dimensionality reduction techniques to optimize component selection. This could be combined with feature selection strategies that combine feature importance scores across multiple classifiers and XAI techniques to ensure consistent identification of valuable features and improve model robustness and reliability.

\section{Conclusion}
 
Our findings reveal that embedding-based feature enrichment offers performance benefits for ensemble classifiers, especially for datasets with limited feature diversity or class representation. These gains can be further amplified with more advanced ensemble classifiers such as XGBoost or CatBoost. For datasets with comprehensive representation, the performance gains are modest at best, highlighting the selective applicability of this approach. Additionally, embedding-enriched models take full advantage of LLM-derived features, which consistently rank among the most influential in feature importance analysis. Hence, these features may play a critical role in uncovering complex data patterns beyond the reach of the original features.

These insights emphasize the potential of contextual embeddings to advance traditional feature engineering, aligning performance improvements with privacy-preserving strategies. We see high potential for research on optimizing embeddings for structured data, refining dimensionality reduction techniques, and exploring additional pre-trained models and embedding dimensions to further enhance generalizability across a broader range of classification tasks.

\bibliographystyle{plain}
\bibliography{literature}

\newpage
\section{Appendix}

\begin{scriptsize}
    
\begin{longtable}{p{1.5cm}|l|l|cp{1.3cm}cc}
\caption{Performance Metrics for Different Feature Subsets Across Datasets and Models}\label{tab:all-results}\\
\toprule
\textbf{Dataset} & \textbf{Model} & \textbf{Feature Subset} & \textbf{Accuracy} & \textbf{Balanced Accuracy} & \textbf{F1 Score} & \textbf{ROC-AUC} \\ 
\midrule
\endfirsthead

\multicolumn{7}{c}{{\bfseries \tablename\ \thetable{} -- continued from previous page}} \\
\toprule
\textbf{Dataset} & \textbf{Model} & \textbf{Feature Subset} & \textbf{Accuracy} & \textbf{Weighted Accuracy} & \textbf{F1 Score} & \textbf{ROC-AUC} \\ 
\midrule
\endhead

\midrule \multicolumn{7}{r}{{Continued on next page}} \\
\midrule
\endfoot

\bottomrule
\endlastfoot

% ---- UCI Covertype ----
\multirow{22}{*}{\parbox{1.5cm}{\centering UCI Covertype}} & RandomForest & Baseline & \textbf{0.9021} & \textbf{0.7828} & \textbf{0.9005} & \textbf{0.9884} \\
 & RandomForest & GPT2\_Selected & 0.5722 & 0.2044 & 0.5310 & 0.7567 \\
 & RandomForest & RoBERTa\_Selected & 0.5947 & 0.2444 & 0.5602 & 0.7952 \\
 & RandomForest & GPT2\_RoBERTa & 0.6067 & 0.2312 & 0.5677 & 0.8316 \\
 & RandomForest & Baseline\_GPT2 & 0.8469 & 0.6835 & 0.8420 & 0.9801 \\
 & RandomForest & Baseline\_RoBERTa & 0.8464 & 0.6770 & 0.8414 & 0.9793 \\
 & RandomForest & Baseline\_GPT2\_RoBERTa\_Selected & 0.8153 & 0.6157 & 0.8084 & 0.9729 \\
 & XGBoost & Baseline & \textbf{0.8603} & \textbf{0.7688} & \textbf{0.8590} & \textbf{0.9831} \\
 & XGBoost & GPT2\_Selected & 0.5766 & 0.2323 & 0.5454 & 0.7884 \\
 & XGBoost & RoBERTa\_Selected & 0.5943 & 0.2639 & 0.5673 & 0.8235 \\
 & XGBoost & GPT2\_RoBERTa & 0.6188 & 0.2984 & 0.5969 & 0.8577 \\
 & XGBoost & Baseline\_GPT2 & 0.8449 & 0.7281 & 0.8429 & 0.9803 \\
 & XGBoost & Baseline\_RoBERTa & 0.8478 & 0.7400 & 0.8459 & 0.9804 \\
 & XGBoost & Baseline\_GPT2\_RoBERTa\_Selected & 0.8399 & 0.7170 & 0.8377 & 0.9789 \\
 & CatBoost & Baseline & \textbf{0.8679} & \textbf{0.7653} & \textbf{0.8664} & \textbf{0.9848} \\
 & CatBoost & GPT2\_Selected & 0.5798 & 0.2391 & 0.5489 & 0.8052 \\
 & CatBoost & RoBERTa\_Selected & 0.6001 & 0.2728 & 0.5734 & 0.8331 \\
 & CatBoost & GPT2\_RoBERTa & 0.6271 & 0.3175 & 0.6054 & 0.8699 \\
 & CatBoost & Baseline\_GPT2 & 0.8498 & 0.7251 & 0.8476 & 0.9812 \\
 & CatBoost & Baseline\_RoBERTa & 0.8522 & 0.7367 & 0.8502 & 0.9818 \\
 & CatBoost & Baseline\_GPT2\_RoBERTa\_Selected & 0.8456 & 0.7169 & 0.8432 & 0.9801 \\
\midrule

% ---- UCI Heart Disease ----
\multirow{22}{*}{\parbox{1.5cm}{\centering UCI Heart Disease}} & RandomForest & Baseline & 0.8167 & 0.8064 & 0.8138 & 0.9175 \\
 & RandomForest & GPT2\_Selected & \textbf{0.8333} & \textbf{0.8317} & \textbf{0.8333} & 0.8917 \\
 & RandomForest & RoBERTa\_Selected & 0.8167 & 0.8165 & 0.8169 & 0.8597 \\
 & RandomForest & GPT2\_RoBERTa & 0.8167 & 0.8165 & 0.8169 & 0.8967 \\
 & RandomForest & Baseline\_GPT2 & 0.8000 & 0.7946 & 0.7991 & 0.9063 \\
 & RandomForest & Baseline\_RoBERTa & 0.8167 & 0.8131 & 0.8163 & \textbf{0.9226} \\
 & RandomForest & Baseline\_GPT2\_RoBERTa\_Selected & \textbf{0.8333} & 0.8283 & 0.8326 & 0.9220 \\
 & XGBoost & Baseline & \textbf{0.8500} & \textbf{0.8468} & \textbf{0.8497} & 0.9080 \\
 & XGBoost & GPT2\_Selected & 0.7500 & 0.7525 & 0.7506 & 0.8799 \\
 & XGBoost & RoBERTa\_Selected & 0.7833 & 0.7828 & 0.7836 & 0.8474 \\
 & XGBoost & GPT2\_RoBERTa & 0.7833 & 0.7795 & 0.7829 & 0.9024 \\
 & XGBoost & Baseline\_GPT2 & 0.8167 & 0.8098 & 0.8153 & \textbf{0.9046} \\
 & XGBoost & Baseline\_RoBERTa & 0.8333 & 0.8283 & 0.8326 & 0.8732 \\
 & XGBoost & Baseline\_GPT2\_RoBERTa\_Selected & 0.8167 & 0.8098 & 0.8153 & 0.8889 \\
 & CatBoost & Baseline & 0.8333 & 0.8283 & 0.8326 & 0.9102 \\
 & CatBoost & GPT2\_Selected & 0.8000 & 0.7980 & 0.8000 & 0.8844 \\
 & CatBoost & RoBERTa\_Selected & 0.7833 & 0.7795 & 0.7829 & 0.8642 \\
 & CatBoost & GPT2\_RoBERTa & \textbf{0.8500} & \textbf{0.8434} & \textbf{0.8488} & 0.9046 \\
 & CatBoost & Baseline\_GPT2 & \textbf{0.8500} & \textbf{0.8434} & \textbf{0.8488} & 0.9136 \\
 & CatBoost & Baseline\_RoBERTa & \textbf{0.8500} & \textbf{0.8434} & \textbf{0.8488} & \textbf{0.9169} \\
 & CatBoost & Baseline\_GPT2\_RoBERTa\_Selected & \textbf{0.8500} & \textbf{0.8434} & \textbf{0.8488} & 0.9158 \\
\midrule

% ---- UCI Letter Recognition ----
\multirow{22}{*}{\parbox{1.5cm}{\centering UCI Letter Recognition}} & RandomForest & Baseline & \textbf{0.9611} &\textbf{ 0.9608} & \textbf{0.9611} & \textbf{0.9993} \\
 & RandomForest & GPT2\_Selected & 0.4237 & 0.4240 & 0.4226 & 0.8766 \\
 & RandomForest & RoBERTa\_Selected & 0.5917 & 0.5914 & 0.5932 & 0.9405 \\
 & RandomForest & GPT2\_RoBERTa & 0.6591 & 0.6586 & 0.6606 & 0.9569 \\
 & RandomForest & Baseline\_GPT2 & 0.9490 & 0.9485 & 0.9490 & 0.9989 \\
 & RandomForest & Baseline\_RoBERTa & 0.9477 & 0.9473 & 0.9478 & 0.9989 \\
 & RandomForest & Baseline\_GPT2\_RoBERTa\_Selected & 0.9394 & 0.9391 & 0.9395 & 0.9988 \\
 & XGBoost & Baseline & \textbf{0.9591} & \textbf{0.9589} & \textbf{0.9592} & \textbf{0.9996} \\
 & XGBoost & GPT2\_Selected & 0.4015 & 0.4018 & 0.4010 & 0.8846 \\
 & XGBoost & RoBERTa\_Selected & 0.5672 & 0.5668 & 0.5696 & 0.9463 \\
 & XGBoost & GPT2\_RoBERTa & 0.6621 & 0.6617 & 0.6635 & 0.9682 \\
 & XGBoost & Baseline\_GPT2 & 0.9424 & 0.9421 & 0.9425 & 0.9992 \\
 & XGBoost & Baseline\_RoBERTa & 0.9427 & 0.9423 & 0.9428 & 0.9992 \\
 & XGBoost & Baseline\_GPT2\_RoBERTa\_Selected & 0.9343 & 0.9340 & 0.9344 & 0.9990 \\
 & CatBoost & Baseline & \textbf{0.9611} & \textbf{0.9608} & \textbf{0.9611} & \textbf{0.9997} \\
 & CatBoost & GPT2\_Selected & 0.4374 & 0.4374 & 0.4362 & 0.9014 \\
 & CatBoost & RoBERTa\_Selected & 0.6010 & 0.6004 & 0.6017 & 0.9555 \\
 & CatBoost & GPT2\_RoBERTa & 0.6995 & 0.6990 & 0.7002 & 0.9752 \\
 & CatBoost & Baseline\_GPT2 & 0.9487 & 0.9484 & 0.9487 & 0.9995 \\
 & CatBoost & Baseline\_RoBERTa & 0.9525 & 0.9522 & 0.9525 & 0.9994 \\
 & CatBoost & Baseline\_GPT2\_RoBERTa\_Selected & 0.9452 & 0.9448 & 0.9452 & 0.9993 \\
\midrule

% ---- UCI Wine Quality ----
\multirow{22}{*}{\parbox{1.5cm}{\centering UCI Wine Quality}} & RandomForest & Baseline & 0.6751 & \textbf{0.3171} & 0.6531 & \textbf{0.8588} \\
 & RandomForest & GPT2\_Selected & 0.5552 & 0.2387 & 0.5227 & 0.6793 \\
 & RandomForest & RoBERTa\_Selected & 0.6151 & 0.2871 & 0.5925 & 0.6964 \\
 & RandomForest & Baseline\_GPT2 & 0.6751 & 0.3138 & 0.6504 & 0.7651 \\
 & RandomForest & Baseline\_RoBERTa & 0.6625 & 0.3030 & 0.6366 & 0.8298 \\
 & RandomForest & GPT2\_RoBERTa & 0.6088 & 0.2660 & 0.5742 & 0.6988 \\
 & RandomForest & Baseline\_GPT2\_RoBERTa\_Selected & \textbf{0.6972} & 0.3136 & \textbf{0.6654} & 0.7384 \\
 & XGBoost & Baseline & 0.6688 & 0.3385 & 0.6544 & \textbf{0.8582} \\
 & XGBoost & GPT2\_Selected & 0.5647 & 0.2546 & 0.5436 & 0.7122 \\
 & XGBoost & RoBERTa\_Selected & 0.5678 & 0.2590 & 0.5485 & 0.7030 \\
 & XGBoost & Baseline\_GPT2 & 0.6593 & \textbf{0.3634} & 0.6468 & 0.7713 \\
 & XGBoost & Baseline\_RoBERTa & 0.6498 & 0.3037 & 0.6308 & 0.8652 \\
 & XGBoost & GPT2\_RoBERTa & 0.5773 & 0.2627 & 0.5552 & 0.7527 \\
 & XGBoost & Baseline\_GPT2\_RoBERTa\_Selected & \textbf{0.7035} & 0.3466 & \textbf{0.6832} & 0.8186 \\
 & CatBoost & Baseline & 0.6940 & \textbf{0.3365} & \textbf{0.6744} & \textbf{0.8639} \\
 & CatBoost & GPT2\_Selected & 0.5741 & 0.2552 & 0.5491 & 0.6735 \\
 & CatBoost & RoBERTa\_Selected & 0.5678 & 0.2655 & 0.5466 & 0.7326 \\
 & CatBoost & Baseline\_GPT2 & 0.6656 & 0.3188 & 0.6459 & 0.8166 \\
 & CatBoost & Baseline\_RoBERTa & \textbf{0.6593} & 0.3107 & 0.6393 & 0.8524 \\
 & CatBoost & GPT2\_RoBERTa & 0.5994 & 0.2805 & 0.5775 & 0.7601 \\
 & CatBoost & Baseline\_GPT2\_RoBERTa\_Selected & \textbf{0.6593} & 0.3073 & 0.6357 & 0.8546 \\
\midrule

% ---- UCI Adult ----
\multirow{22}{*}{UCI Adult} & RandomForest & Baseline & 0.8560 & \textbf{0.7765} & 0.8518 & 0.9089 \\
 & RandomForest & GPT2\_Selected & 0.8291 & 0.7157 & 0.8184 & 0.8702 \\
 & RandomForest & RoBERTa\_Selected & 0.8239 & 0.7070 & 0.8124 & 0.8693 \\
 & RandomForest & Baseline\_GPT2 & 0.8594 & 0.7686 & 0.8529 & 0.9119 \\
 & RandomForest & Baseline\_RoBERTa & \textbf{0.8604} & 0.7690 & \textbf{0.8538} & \textbf{0.9112} \\
 & RandomForest & GPT2\_RoBERTa & 0.8347 & 0.7240 & 0.8244 & 0.8806 \\
 & RandomForest & Baseline\_GPT2\_RoBERTa\_Selected & 0.8593 & 0.7641 & 0.8519 & 0.9095 \\
 & XGBoost & Baseline & \textbf{0.8752} & \textbf{0.8027} & \textbf{0.8716} & \textbf{0.9320} \\
 & XGBoost & GPT2\_Selected & 0.8304 & 0.7309 & 0.8231 & 0.8688 \\
 & XGBoost & RoBERTa\_Selected & 0.8226 & 0.7230 & 0.8156 & 0.8676 \\
 & XGBoost & Baseline\_GPT2 & 0.8727 & 0.8011 & 0.8693 & 0.9282 \\
 & XGBoost & Baseline\_RoBERTa & 0.8719 & 0.8009 & 0.8687 & 0.9290 \\
 & XGBoost & GPT2\_RoBERTa & 0.8299 & 0.7389 & 0.8246 & 0.8770 \\
 & XGBoost & Baseline\_GPT2\_RoBERTa\_Selected & 0.8659 & 0.7905 & 0.8621 & 0.9256 \\
 & CatBoost & Baseline & \textbf{0.8753} & \textbf{0.8024} & \textbf{0.8716} & \textbf{0.9327} \\
 & CatBoost & GPT2\_Selected & 0.8378 & 0.7387 & 0.8303 & 0.8775 \\
 & CatBoost & RoBERTa\_Selected & 0.8320 & 0.7289 & 0.8238 & 0.8760 \\
 & CatBoost & Baseline\_GPT2 & 0.8733 & 0.7966 & 0.8691 & 0.9310 \\
 & CatBoost & Baseline\_RoBERTa & 0.8733 & 0.7975 & 0.8692 & 0.9307 \\
 & CatBoost & GPT2\_RoBERTa & 0.8377 & 0.7447 & 0.8316 & 0.8867 \\
 & CatBoost & Baseline\_GPT2\_RoBERTa\_Selected & 0.8724 & 0.7957 & 0.8682 & 0.9302 \\
\midrule

% ---- Car Evaluation ----
\multirow{22}{*}{\parbox{1.5cm}{\centering Car Evaluation}} & RandomForest & Baseline & \textbf{0.9064} & \textbf{0.7784} & \textbf{0.9030} & 0.9800 \\
 & RandomForest & GPT2\_Selected & 0.7749 & 0.4937 & 0.7411 & 0.9313 \\
 & RandomForest & RoBERTa\_Selected & 0.8099 & 0.5356 & 0.7793 & 0.9525 \\
 & RandomForest & Baseline\_GPT2\_Selected & 0.8743 & 0.6762 & 0.8655 & 0.9822 \\
 & RandomForest & Baseline\_RoBERTa\_Selected & 0.8830 & 0.7021 & 0.8750 & 0.9849 \\
 & RandomForest & GPT2\_RoBERTa\_Selected & 0.8275 & 0.5668 & 0.8019 & 0.9641 \\
 & RandomForest & Baseline\_GPT2\_RoBERTa\_Selected & 0.8655 & 0.6641 & 0.8534 & \textbf{0.9861} \\
 & XGBoost & Baseline & \textbf{0.9883} & \textbf{0.9572} & \textbf{0.9882} & \textbf{0.9983} \\
 & XGBoost & GPT2\_Selected & 0.7895 & 0.5804 & 0.7776 & 0.9244 \\
 & XGBoost & RoBERTa\_Selected & 0.8333 & 0.5732 & 0.8182 & 0.9525 \\
 & XGBoost & Baseline\_GPT2\_Selected & 0.9240 & 0.7890 & 0.9249 & 0.9904 \\
 & XGBoost & Baseline\_RoBERTa\_Selected & 0.9211 & 0.7630 & 0.9193 & 0.9879 \\
 & XGBoost & GPT2\_RoBERTa\_Selected & 0.8684 & 0.6423 & 0.8609 & 0.9719 \\
 & XGBoost & Baseline\_GPT2\_RoBERTa\_Selected & 0.9211 & 0.7493 & 0.9207 & 0.9888 \\
 & CatBoost & Baseline & \textbf{0.9795} & \textbf{0.9496} & \textbf{0.9794} & \textbf{0.9974} \\
 & CatBoost & GPT2\_Selected & 0.8158 & 0.6192 & 0.8057 & 0.9409 \\
 & CatBoost & RoBERTa\_Selected & 0.8450 & 0.6092 & 0.8291 & 0.9604 \\
 & CatBoost & Baseline\_GPT2\_Selected & 0.9415 & 0.8339 & 0.9420 & 0.9929 \\
 & CatBoost & Baseline\_RoBERTa\_Selected & 0.9415 & 0.7952 & 0.9395 & 0.9903 \\
 & CatBoost & GPT2\_RoBERTa\_Selected & 0.8918 & 0.7052 & 0.8847 & 0.9803 \\
 & CatBoost & Baseline\_GPT2\_RoBERTa\_Selected & 0.9240 & 0.7959 & 0.9237 & 0.9916 \\
\midrule

% ---- Pima Indians Diabetes ----
\multirow{22}{*}{\parbox{1.5cm}{\centering Pima Indians Diabetes}} & RandomForest & Baseline & \textbf{0.7632} & \textbf{0.7130} & \textbf{0.7548} & \textbf{0.8408} \\
 & RandomForest & GPT2\_Selected & 0.7039 & 0.6237 & 0.6783 & 0.7252 \\
 & RandomForest & RoBERTa\_Selected & 0.7105 & 0.6375 & 0.6901 & 0.6772 \\
 & RandomForest & Baseline\_GPT2 & 0.7171 & 0.6426 & 0.6957 & 0.8334 \\
 & RandomForest & Baseline\_RoBERTa & 0.7566 & 0.6816 & 0.7355 & 0.8345 \\
 & RandomForest & GPT2\_RoBERTa & 0.6711 & 0.5809 & 0.6367 & 0.7300 \\
 & RandomForest & Baseline\_GPT2\_RoBERTa\_Selected & 0.7368 & 0.6621 & 0.7155 & 0.8174 \\
 & XGBoost & Baseline & 0.7237 & 0.6827 & 0.7195 & 0.8094 \\
 & XGBoost & GPT2\_Selected & 0.7039 & 0.6544 & 0.6966 & 0.7263 \\
 & XGBoost & RoBERTa\_Selected & 0.6382 & 0.5688 & 0.6211 & 0.6387 \\
 & XGBoost & Baseline\_GPT2 & 0.7237 & 0.6783 & 0.7178 & 0.8010 \\
 & XGBoost & Baseline\_RoBERTa & 0.7171 & 0.6557 & 0.7038 & 0.8140 \\
 & XGBoost & GPT2\_RoBERTa & 0.6447 & 0.5739 & 0.6264 & 0.7134 \\
 & XGBoost & Baseline\_GPT2\_RoBERTa\_Selected & \textbf{0.7434} & \textbf{0.6891} & \textbf{0.7334} & \textbf{0.8311} \\
 & CatBoost & Baseline & \textbf{0.7632} & \textbf{0.7130} & \textbf{0.7548} & \textbf{0.8557} \\
 & CatBoost & GPT2\_Selected & 0.6974 & 0.6230 & 0.6761 & 0.7305 \\
 & CatBoost & RoBERTa\_Selected & 0.6842 & 0.6085 & 0.6620 & 0.6770 \\
 & CatBoost & Baseline\_GPT2 & 0.7632 & \textbf{0.7130} & \textbf{0.7548} & 0.8306 \\
 & CatBoost & Baseline\_RoBERTa & 0.7566 & 0.6992 & 0.7451 & 0.8458 \\
 & CatBoost & GPT2\_RoBERTa & 0.6711 & 0.5984 & 0.6511 & 0.7374 \\
 & CatBoost & Baseline\_GPT2\_RoBERTa\_Selected & 0.7566 & 0.6992 & 0.7451 & 0.8256 \\
\midrule

% ---- Titanic ----
\multirow{22}{*}{Titanic} & RandomForest & Baseline & \textbf{0.8531} & \textbf{0.8392} & \textbf{0.8522} & \textbf{0.8855} \\
 & RandomForest & GPT2\_Selected & 0.7966 & 0.7823 & 0.7960 & 0.8528 \\
 & RandomForest & RoBERTa\_Selected & 0.8023 & 0.7841 & 0.8007 & 0.8774 \\
 & RandomForest & Baseline\_GPT2 & 0.8475 & 0.8264 & 0.8451 & 0.8741 \\
 & RandomForest & Baseline\_RoBERTa & 0.8362 & 0.8283 & 0.8364 & 0.8853 \\
 & RandomForest & GPT2\_RoBERTa & 0.8192 & 0.8062 & 0.8187 & 0.8610 \\
 & RandomForest & Baseline\_GPT2\_RoBERTa\_Selected & 0.8305 & 0.8098 & 0.8282 & 0.8741 \\
 & XGBoost & Baseline & \textbf{0.8475} & \textbf{0.8347} & \textbf{0.8468} & 0.8716 \\
 & XGBoost & GPT2\_Selected & 0.8192 & 0.8007 & 0.8175 & 0.8548 \\
 & XGBoost & RoBERTa\_Selected & 0.8079 & 0.7998 & 0.8084 & 0.8601 \\
 & XGBoost & GPT2\_RoBERTa & 0.8249 & 0.8108 & 0.8241 & 0.8621 \\
 & XGBoost & Baseline\_GPT2 & 0.8249 & 0.8246 & 0.8262 & 0.8736 \\
 & XGBoost & Baseline\_RoBERTa & 0.8249 & 0.8135 & 0.8246 & \textbf{0.8739} \\
 & XGBoost & GPT2\_RoBERTa & 0.8249 & 0.8163 & 0.8251 & 0.8587 \\
 & CatBoost & Baseline & 0.8249 & 0.7997 & 0.8214 & 0.8837 \\
 & CatBoost & GPT2\_Selected & 0.8305 & 0.8126 & 0.8289 & 0.8559 \\
 & CatBoost & RoBERTa\_Selected & 0.8249 & 0.8163 & 0.8251 & 0.8826 \\
 & CatBoost & Baseline\_GPT2 & \textbf{0.8475} & 0.8291 & \textbf{0.8457} & 0.8627 \\
 & CatBoost & Baseline\_RoBERTa & 0.8418 & \textbf{0.8328} & 0.8418 & \textbf{0.8944} \\
 & CatBoost & GPT2\_RoBERTa & 0.8362 & 0.8255 & 0.8359 & 0.8753 \\
 & CatBoost & Baseline\_GPT2\_RoBERTa\_Selected & 0.8418 & 0.8301 & 0.8413 & 0.8797 \\

\bottomrule
\end{longtable}
\end{scriptsize}

\begin{figure}[h!]
    \centering
    \makebox[\textwidth]{ 
    \resizebox{1.3\textwidth}{!}{
    \begin{minipage}{0.5\textwidth}
        \centering
        \includegraphics[width=\linewidth, height=6.5cm]{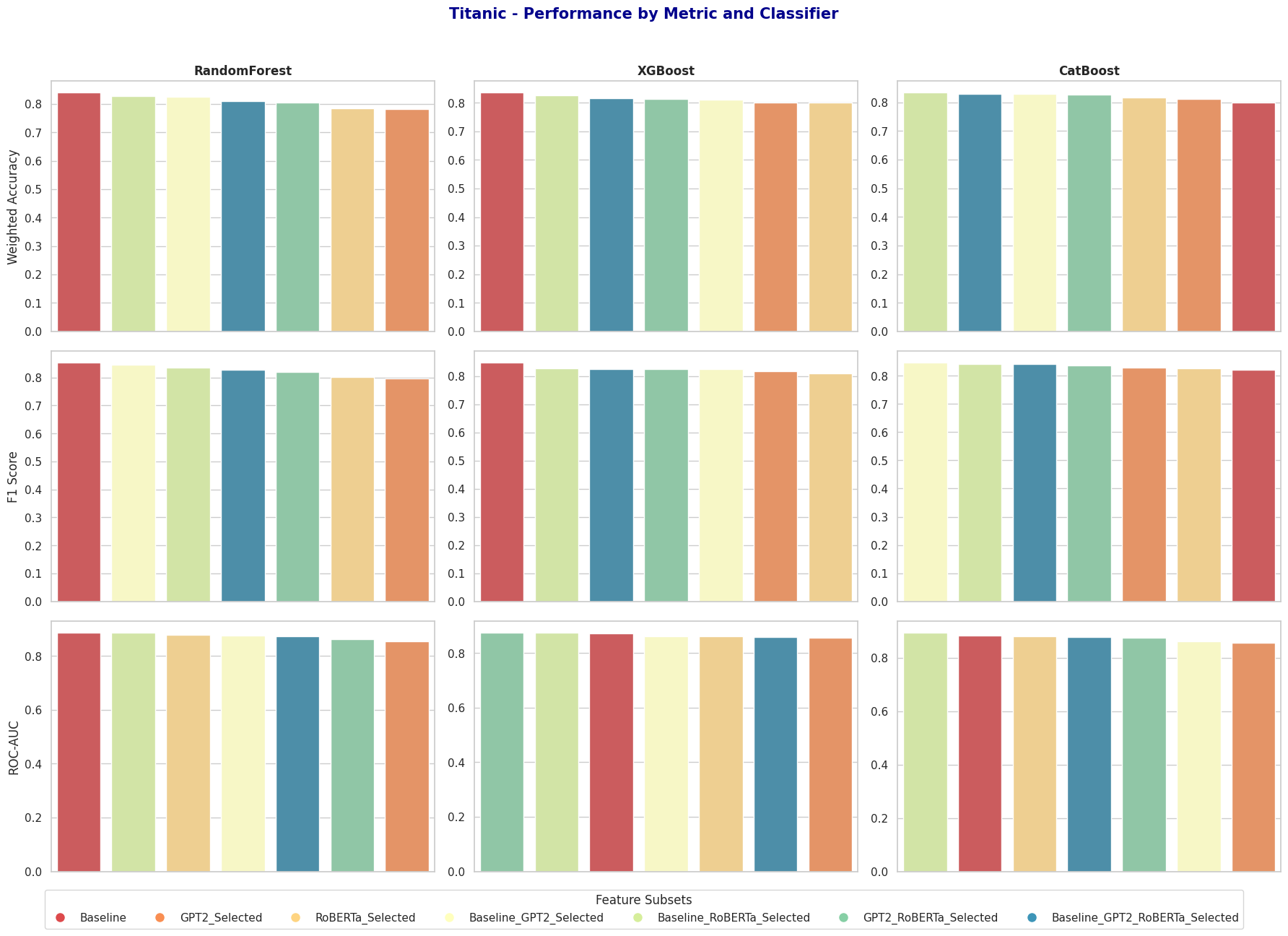}
    \end{minipage}\hfill
    \begin{minipage}{0.5\textwidth}
        \centering
        \includegraphics[width=\linewidth, height=6.5cm]{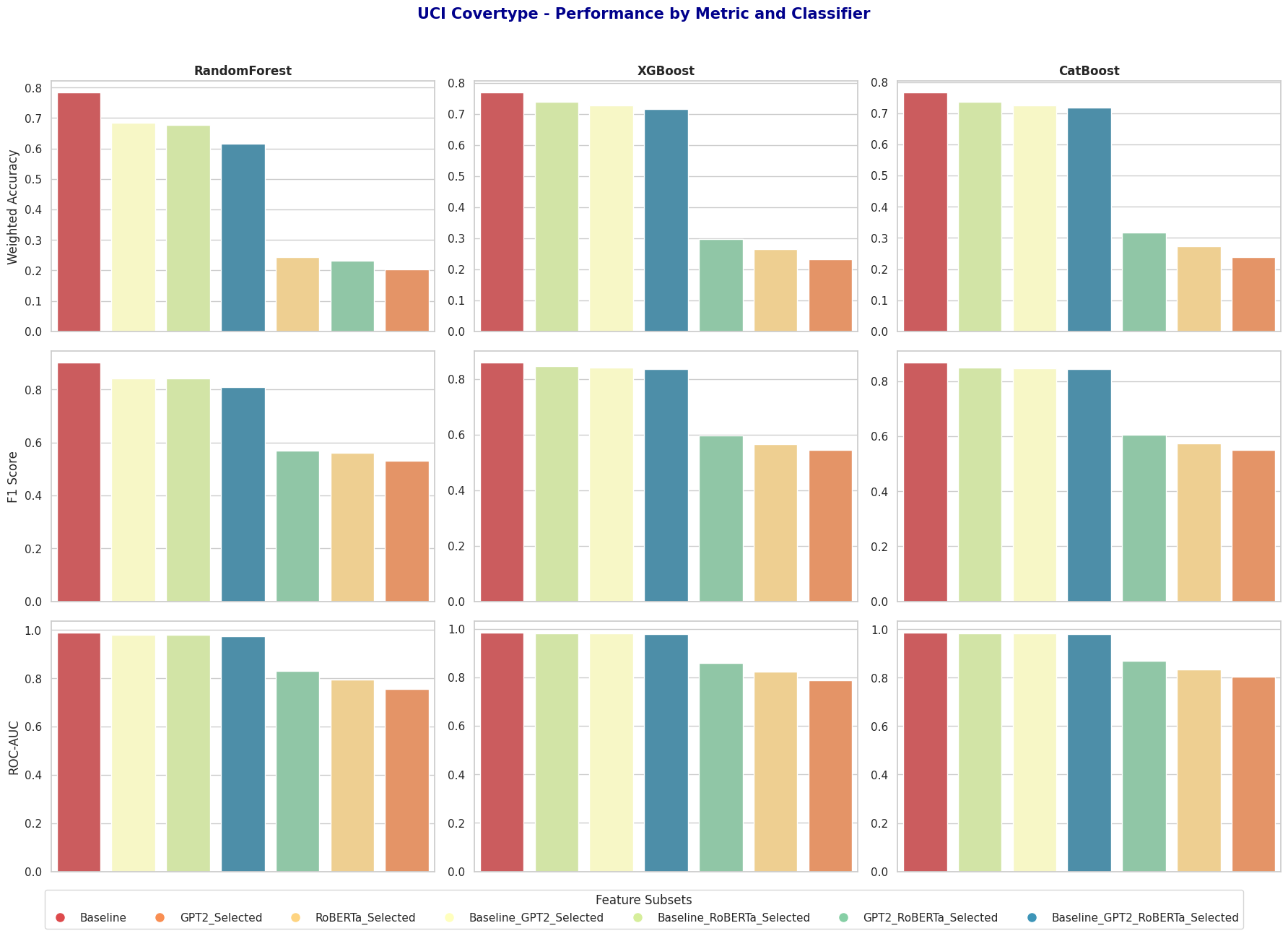}
    \end{minipage}
    }
    }
    \makebox[\textwidth]{ 
    \resizebox{1.3\textwidth}{!}{
    \begin{minipage}{0.5\textwidth}
        \centering
        \includegraphics[width=\linewidth, height=6.5cm]{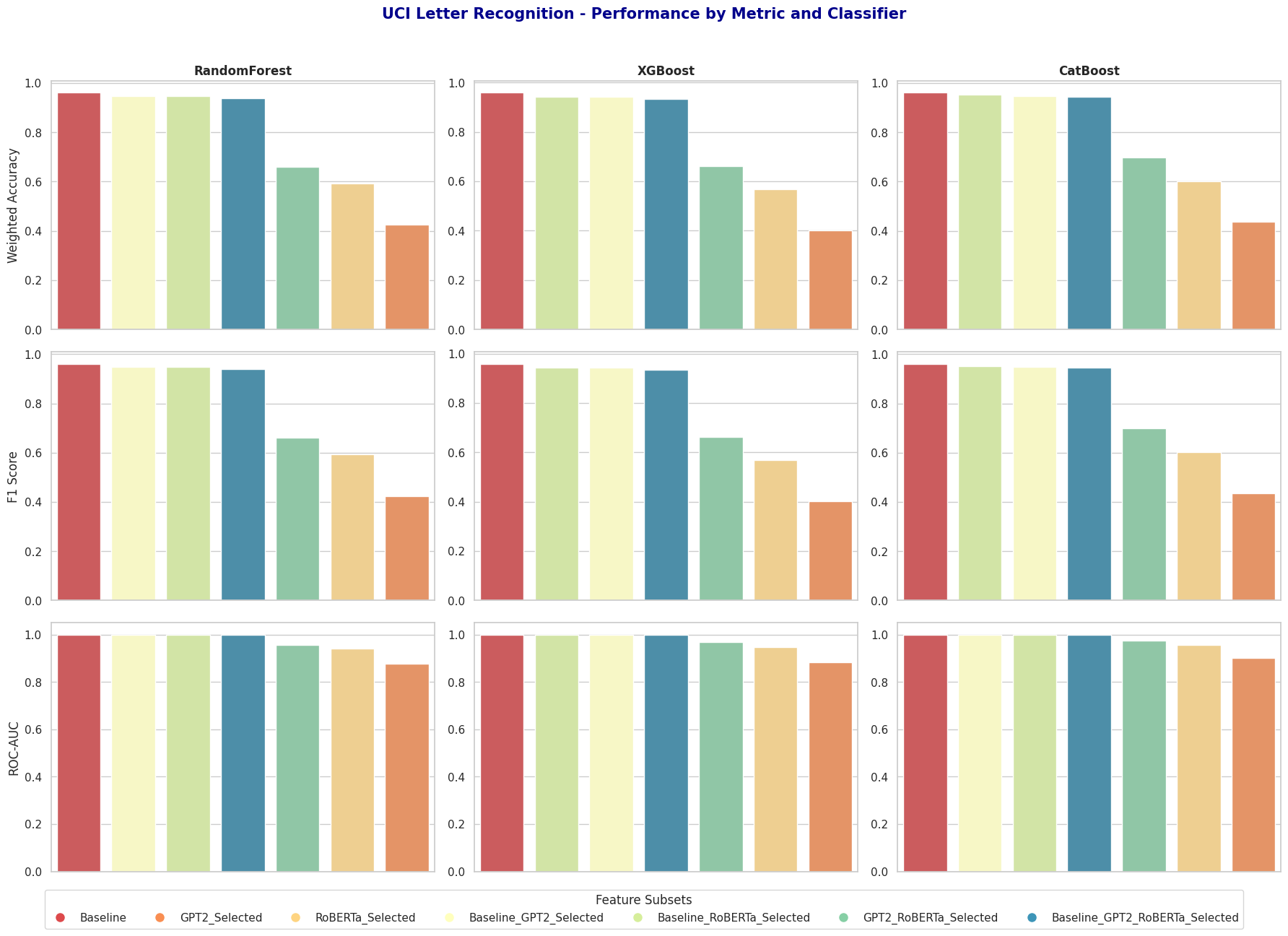}
    \end{minipage}\hfill
    \begin{minipage}{0.5\textwidth}
        \centering
        \includegraphics[width=\linewidth, height=6.5cm]{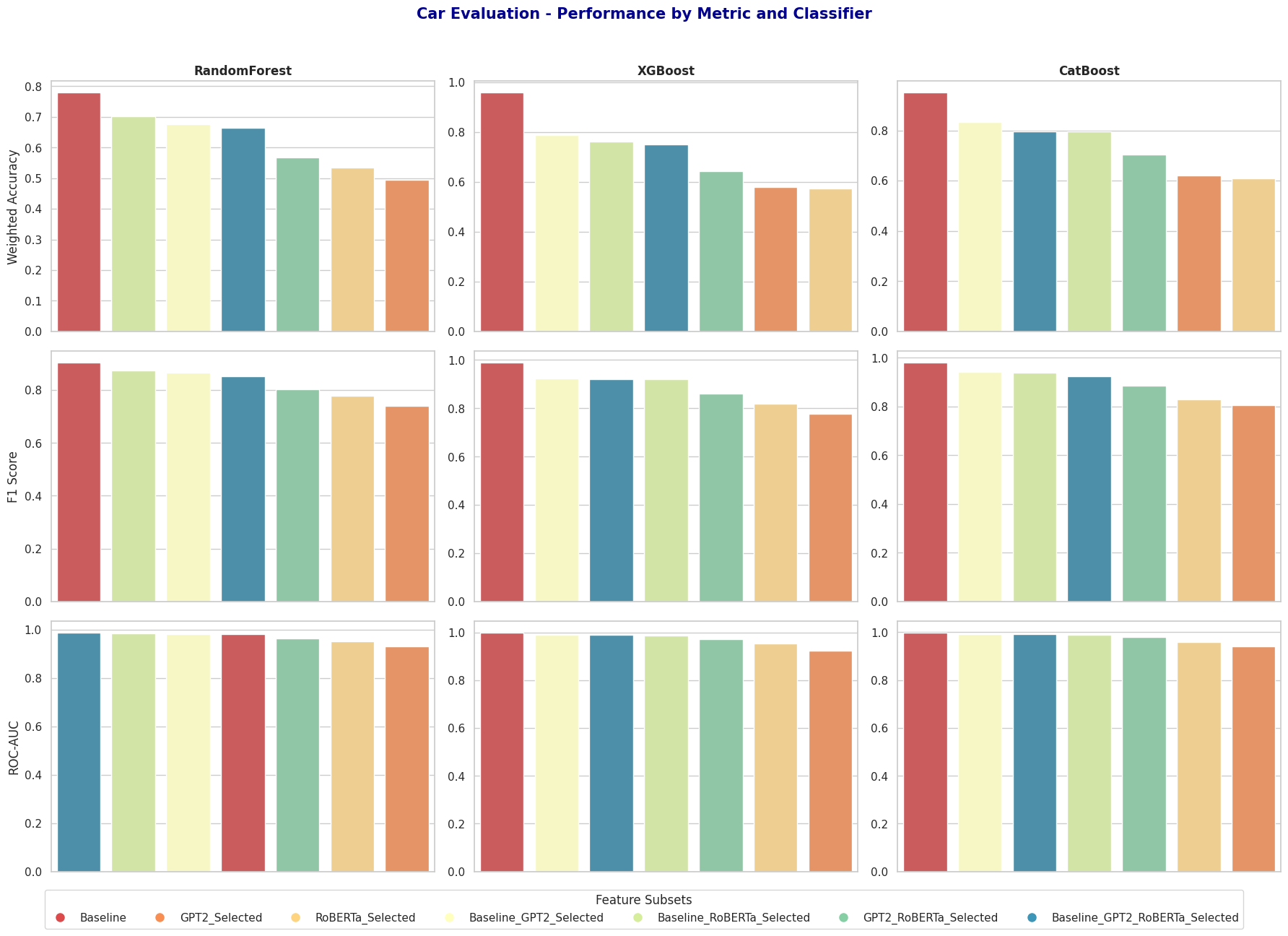}
    \end{minipage}
    }
    }
    \caption{Performance metrics (Weighted/Balanced Accuracy, F1 Score, and ROC-AUC) for different feature subsets on multiple datasets (\texttt{Titanic}, \texttt{UCI Covertype}, \texttt{UCI Letter Recgonition}, and \texttt{Car Evaluation}) using Random Forest, XGBoost, and CatBoost classifiers. The plots show how different feature subsets influence classifier performance, with results sorted by feature subset to facilitate comparisons.}
\label{fig:combined_feature_performance_plots_2}
\end{figure}

\end{document}